\documentclass[10pt,journal,compsoc]{IEEEtran}

\ifCLASSOPTIONcompsoc
  \usepackage[nocompress]{cite}
\else
  \usepackage{cite}
\fi

\usepackage{url}
\usepackage{times}
\usepackage{epsfig}
\usepackage{graphicx}
\usepackage{amsmath}
\usepackage{amssymb}
\usepackage[numbers,sort]{natbib} 
\usepackage{float}
\usepackage{subcaption}
\usepackage{tabularx}
\usepackage{booktabs}
\usepackage{colortbl}
\usepackage{xcolor}
\definecolor{lightgray}{gray}{0.8}
\definecolor{caner}{rgb}{0, 0, 0}

\begin{document}

\newcommand{\eg}{\MakeLowercase{\textit{e.g.~}}}
\newcommand{\ie}{\MakeLowercase{\textit{i.e.~}}}
\newcommand{\etc}{\MakeLowercase{\textit{etc.~}}}
\newcommand{\wrt}{\MakeLowercase{\textit{w.r.t.~}}}
\newcommand{\etal}{\MakeLowercase{\textit{et~al.~}}}

%
\title{Towards Measuring Fairness in AI:\\the Casual Conversations Dataset}

\author{Caner~Hazirbas,
        Joanna~Bitton,
        Brian~Dolhansky,
        Jacqueline~Pan,
        Albert~Gordo,
        and~Cristian~Canton~Ferrer
\IEEEcompsocitemizethanks{\IEEEcompsocthanksitem C. Hazirbas (hazirbas@fb.com), J. Bitton, J. Pan, A. Gordo and C. Canton Ferrer are with Facebook AI. B. Dolhansky was with Facebook AI.}%
}


\IEEEtitleabstractindextext{%
\begin{abstract}
This paper introduces a novel dataset to help researchers evaluate their computer vision and audio models for accuracy across a diverse set of age, genders, apparent skin tones and ambient lighting conditions. Our dataset is composed of 3,011 subjects and contains over 45,000 videos, with an average of 15 videos per person. The videos were recorded in multiple U.S. states with a diverse set of adults in various age, gender and apparent skin tone groups. A key feature is that each subject agreed to participate for their likenesses to be used. Additionally, our age and gender annotations are provided by the subjects themselves. A group of trained annotators labeled the subjects' apparent skin tone using the Fitzpatrick skin type scale. Moreover, annotations for videos recorded in low ambient lighting are also provided. As an application to measure robustness of predictions across certain attributes, we provide a comprehensive study on the top five winners of the DeepFake Detection Challenge (DFDC). Experimental evaluation shows that the winning models are less performant on some specific groups of people, such as subjects with darker skin tones and thus may not generalize to all people. In addition, we also evaluate the state-of-the-art apparent age and gender classification methods. Our experiments provides a thorough analysis on these models in terms of fair treatment of people from various backgrounds.
\end{abstract}

}

\maketitle

\IEEEdisplaynontitleabstractindextext
\IEEEpeerreviewmaketitle

\IEEEraisesectionheading{\section{Introduction}\label{sec:introduction}}
Fairness in AI is an emerging topic in computer vision~\cite{Kaiyu20,Vries2019} and has proven  indispensable to develop unbiased AI models that are fair and inclusive to individuals from any background. Recent studies~\cite{Eidinger14,Kaerkkaeinen19,Zhifei17} suggest that top performing AI models trained on datasets that are created without considering fair distribution across sub-groups and thus quite unbalanced, do not necessarily reflect the outcome in real world. On the contrary, they may perform poorly and may be biased towards certain groups of people.

\begin{figure}[ht]
    \centering
    \includegraphics[width=\linewidth]{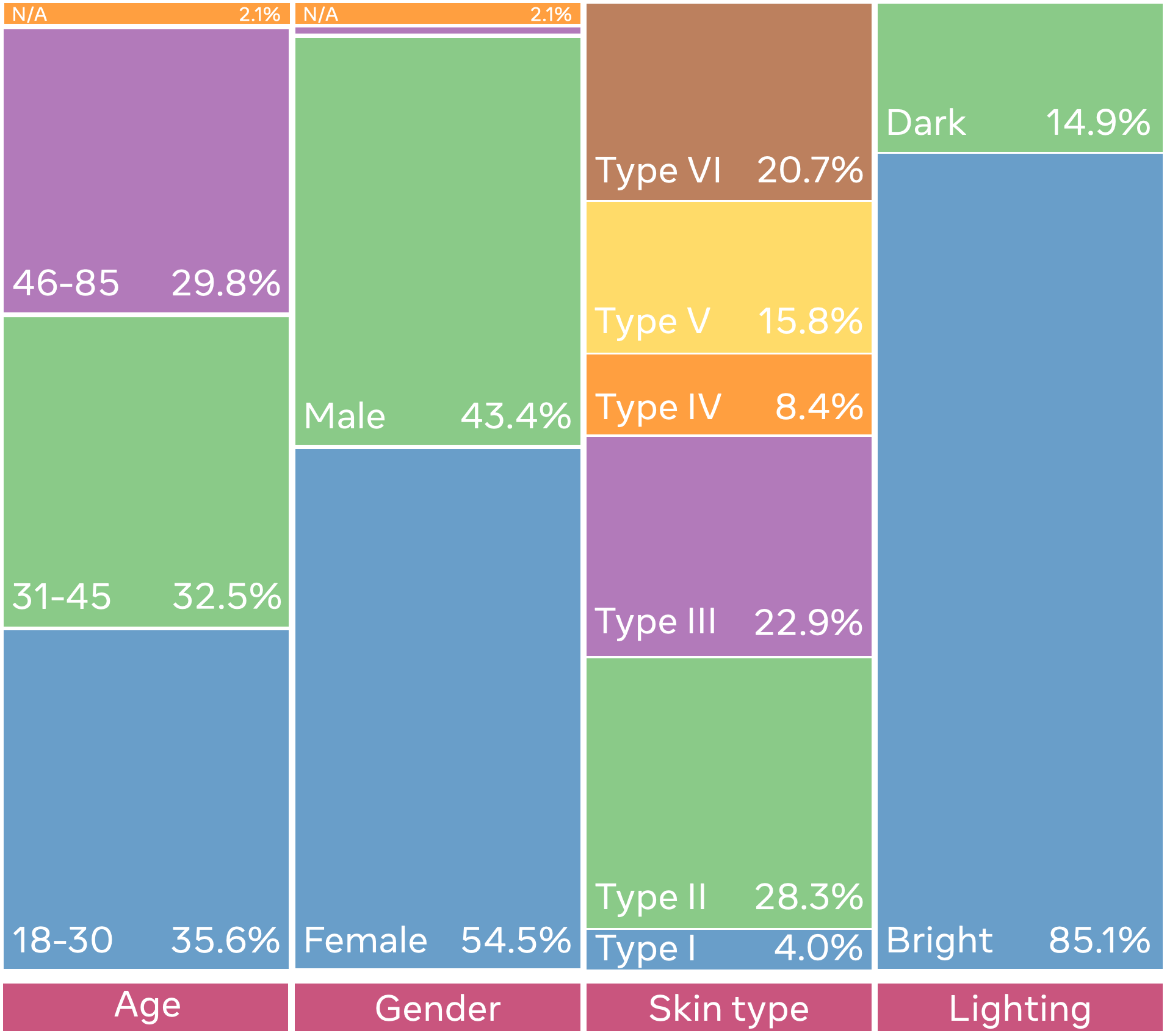}
    \caption{\label{fig:dist}\textbf{Casual Conversations dataset} per-category distributions. Age and gender distributions are pretty balanced. Only 0.1\% of participants identified themselves as \textit{Other} gender (purple bar in the ``Gender'' column). Consecutive pairs of skin types can be grouped into three sub-categories for a uniform distribution. To balance lighting, we also provide sub-sampled dataset consists of two videos per actor, where one is \textit{Dark} when possible.}
\end{figure}

\begin{figure*}
    \centering
    \includegraphics[width=\textwidth]{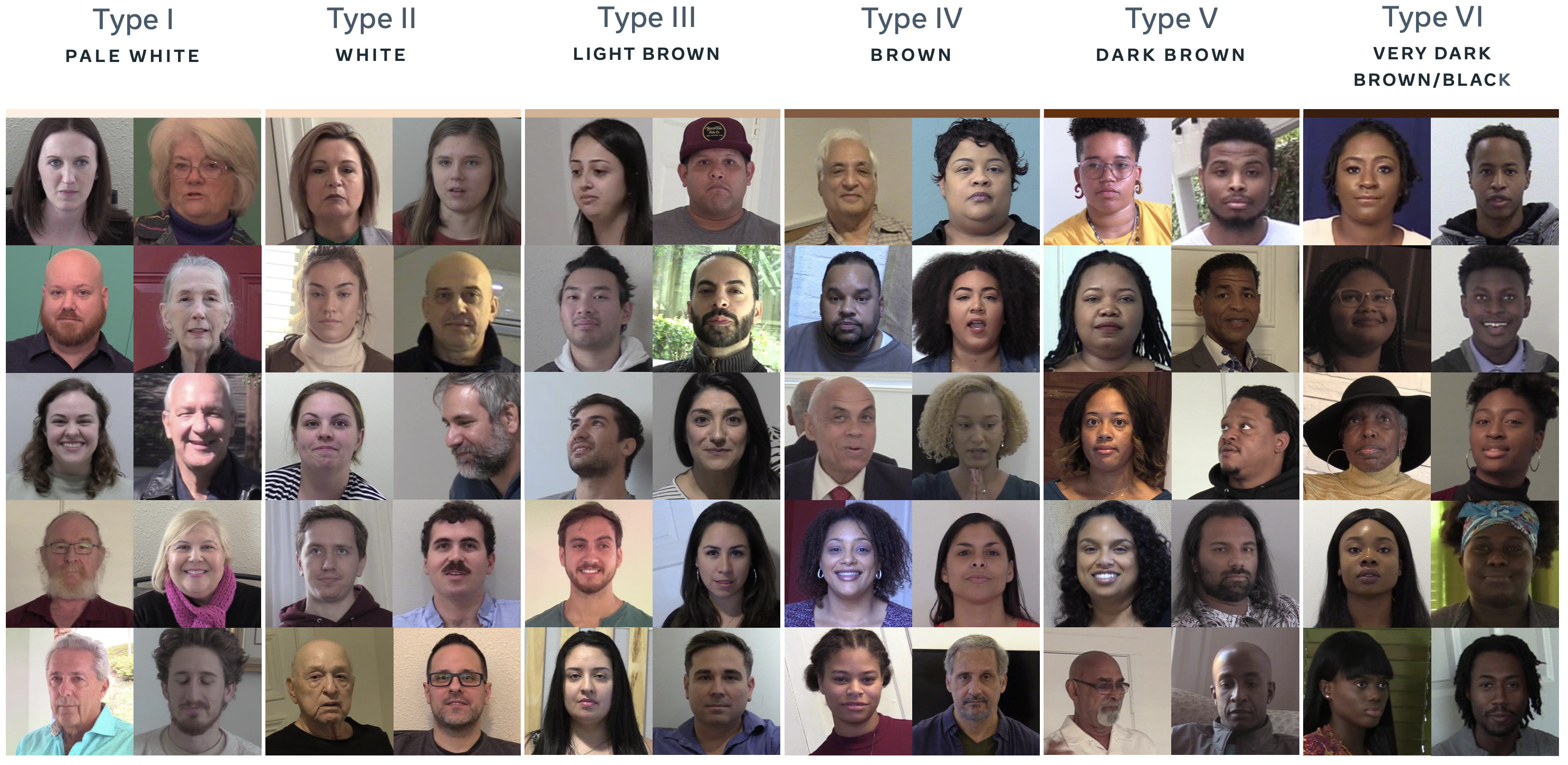}
    \caption{\label{fig:visual1_hero_still}\textbf{Example face crops} from the \textit{Casual Conversations} dataset, categorized by their apparent Fitzpatrick skin types.}
\end{figure*}

Deepfake detectors are able to differentiate between real and fake videos by accumulating classifier responses on individual frames. Although aggregating per-frame predictions removes most outliers for a robust and accurate classification, bias in face detectors may change the final result drastically and cause deepfake detectors to fail. Are these detectors, e.g. \cite{King09, Zhang16, Seferbekov2020, WM2020, NTechLab2020, Eighteen2020, Medics2020}, capable of detecting faces from various age groups, genders or skin tones? If so, how often do deepfake detectors fail based on this reason? Is there a way we can measure the vulnerabilities of deepfake detectors?

To address the aforementioned concerns, we propose a dataset composed of video recordings containing 3,011 individuals with a diverse set of age, genders and apparent skin types. Participants, who were paid actors gave their permission for their likeness to be used for improving AI, in the video recordings casually speak about various topics and sometimes depict a range of facial expressions. Thus, we call the dataset \textit{Casual Conversations}. The dataset includes a unique identifier and age, gender, apparent skin type annotations for each subject. A distinguishing feature of our dataset is that age and gender annotations are provided by the subjects themselves. We prefer this human-centered approach and believe it allows our data to have a relatively unbiased view of age and gender. As a third dimension in our dataset, we annotated the apparent skin tone of each subject using the Fitzpatrick~\cite{FitzPatrick75} scale; we also label videos recorded in low ambient lighting. This set of attributes allow us to measure model robustness on four dimensions: age, gender, apparent skin tone and ambient lighting.

Although \textit{Casual Conversations} is intended to evaluate robustness of AI models across several facial attributes, we believe that its value is greater and indispensable for many other open challenges. Image inpainting, developing temporally consistent models, audio understanding, responsible AI on facial attribute classification and handling low-light scenarios in the aforementioned problems are potential application areas of this dataset.

We organize the paper as follows;~Section~\ref{sec:relatedwork} provides a comprehensive background on fairness in AI, up-to-date facial attribute datasets, deepfake detection and current challenges in personal attribute classification.~Section~\ref{sec:dataset} describes the data acquisition process and the annotation pipeline for our dataset. ~Section~\ref{sec:experiments} analyzes the biases of top five DFDC winners as well as the state-of-the-art apparent age and gender classification models, using our dataset. Consequently, we finalize our findings and provide an overview of the results in~Section~\ref{sec:conclusion}.

\section{\label{sec:relatedwork}Related Work}
\noindent\textbf{Fairness in AI} challenges the field of artificial intelligence to be more inclusive, fair and responsible. Research has clearly shown that deep networks that achieve a high performance on certain datasets are likely to favor only sub-groups of people due to the imbalanced distribution of the categories in the data~\cite{Larrazabal20}. Buolamwini and Gebru~\cite{Buolamwini18} pointed out that the IJB-A~\cite{Klare15} and Adience~\cite{Eidinger14} datasets are composed of mostly lighter skin toned subjects. Raji~\etal~\cite{Raji19,Raji20} analyze the commercial impact Gender Shades~\cite{Buolamwini18} and discuss ethical concerns auditing facial processing technologies. Du~\etal~\cite{Du20} provide a comprehensive review on recent developments of fairness in deep learning and discuss potential fairness mitigation approaches in deep learning.  Meanwhile, Barocas~\etal~\cite{Barocas19} are in the process of compiling a book that intends to give a fresh perspective on machine learning in regards to fairness as a central concern and discusses possible mitigation strategies on the ethical challenges in AI.

\textcolor{caner}{
Several other recent studies are carried out to find the fairness gap in biometric technologies. Grother~\etal~\cite{Grother2018} evaluated demographic effects of the face recognition technologies on four large datasets of photos collected in U.S. governmental applications. They showed that majority of the contemporary face recognition algorithms exhibit large demographic differentials. In following,~\cite{Cook2019} examined eleven commercial face biometrics systems for the effect of demographic factors and found out that both speed and the accuracy of these systems were influenced by demographic factors. Howard~\etal~\cite{Howard2019} developed a framework to identify biometric performance differentials in terms of false positive and negative outcomes and showed that majority of previous works focused on false negatives.~\cite{Drozdowski2020} provided a detailed overview on the algorithmic bias in terms of biometrics and carried out a survey on literature that focus on biometric bias estimation and mitigation. In~\cite{Krishnapriya2020}, Krishnapriya~\etal explored issues in face recognition accuracy discrepancy based on race and skin tone and showed that cohort of African-American has a higher false match rate. Moreover,~\cite{Albiero2020} compared accuracy of face recognition for ``man'' and ``woman'' genders and showed that regardless of facial expression, head pose and forehead occlusion, accuracy discrepancy between ``man'' and ``woman'' was persistent and trained models (even on a balanced dataset) were less accurate for ``woman'' gender.
}\\

\noindent\textbf{Facial attribute datasets}~\cite{Eidinger14,Kaerkkaeinen19,Zhifei17} are created to train and validate face recognition, age, and gender classification models. However, provided facial attributes in these datasets are hand-labelled and annotated by third-parties. Although it has been claimed that the annotations are uniformly distributed over different attributes,~\eg age and gender, there is no guarantee on the accuracy of these annotations. An individual's visual appearance may differ significantly from their own self-identification which will thus result as bias in the dataset. In contrast, we provide age and gender annotations that are \textit{self-identified} by the subjects.~\textcolor{caner}{Non-commercial MORPH dataset~\cite{Ricanek2006} is composed of longitudinal face images of people and provides binary gender labels come from government-furnished ID. Although, this dataset is collected with legal considerations and IRB approval, it is very imbalanced for binary genders\footnote{~\url{http://people.uncw.edu/vetterr/MORPH-NonCommercial-Stats.pdf}} (only 15.4\% ``female'')}. Aside from age and gender, public benchmarks tend to also provide annotated ethnicity labels. However, we find that labeling the ethnicity of subjects could lead to inaccuracies. Raters may have unconscious biases towards certain ethnic groups that may reduce the labelling accuracy in the provided annotations~\textcolor{caner}{\cite{Otoole1996}}. In the FairFace~\cite{Kaerkkaeinen19} dataset paper, the authors claim that skin tone is a one dimensional concept in comparison to ethnicity because lighting is a big factor when deciding on the skin tone as a subject's skin tone may vary over time. Although these claims sound reasonable, the ethnicity attribute is still \textcolor{caner}{subjective} and can conceptually cause confusions in many aspects; for example, there may be no  difference in facial appearance of African-American and African people, although, they may be referred to with two distinct racial categories. We, therefore, have opted to annotate the apparent skin tone of each subject. Our dataset is composed of multiple recordings (~\ie on avg. 15) per actor, so annotators voted based on the sampled frames of these videos. Since these videos were captured in varying ambient lighting conditions, we alleviate the aforementioned concerns stated in~\cite{Kaerkkaeinen19}.~\textcolor{caner}{Nevertheless, color-correction may still be necessary for improved inter-rater agreement~\cite{Krishnapriya2021}.}\\

\noindent\textbf{Deepfake detection} in media forensics is a burgeoning field in machine learning that attempts to classify realistic artificially-generated videos. Fake content generation has rapidly evolved in the past several years and recent studies~\cite{Nyugen19} show that the gap between the deepfake generators and detectors has been growing quickly. The DeepFake Detection Challenge (DFDC) ~\cite{DFDCPreview, DFDC2020} provided researchers the opportunity to develop state-of-the-art detectors to tackle fake videos. However, there are still open questions to address such as the robustness of these detectors across various age, gender, apparent skin tone groups and ambience lighting conditions. While detectors rely on face detection methods that are used by a majority of researchers, there is still no clear understanding as to how accurately these face detectors perform on various subgroups of people, such as specific genders, darker skin tones, younger people,~\etc For this reason, we believe that the \textit{Casual Conversations} dataset will provide a valuable tool to measure robustness of state-of-the-art deepfake detection approaches.\\

\noindent\textbf{Apparent age and gender classification} has been a rapidly growing research field over a decade but recently took more attention after tremendous increase in social media usage. Therefore, apparent age and gender prediction is still an active research field investigated in automated human biometrics and facial attribute classification methods. Levi \& Hassner~\cite{LeviHassner15} proposed an end-to-end trained Convolutional Neural Network (CNN) on Adience benchmark~\cite{Eidinger14} to predict apparent age and gender from faces. Lee~\etal~\cite{Lee2018} further developed a system for mobiles and proposed an efficient, lightweight multi-task CNN for simultaneous apparent age and gender classification. Serengil~\etal~\cite{Serengil2020} recently presented a hybrid face recognition framework composed of the state-of-the-art face recognition approaches. Nevertheless, none of these models were evaluated against apparent skin type variations and ambient lighting conditions. Therefore, we present a close look into the results of these methods on our dataset to measure robustness of the recent face technologies.

\newcommand\scale{0.48}
\begin{figure*}
    \begin{tabular}{cc}
    \begin{subfigure}{\scale\textwidth}
        \includegraphics[width=\linewidth, height=3cm]{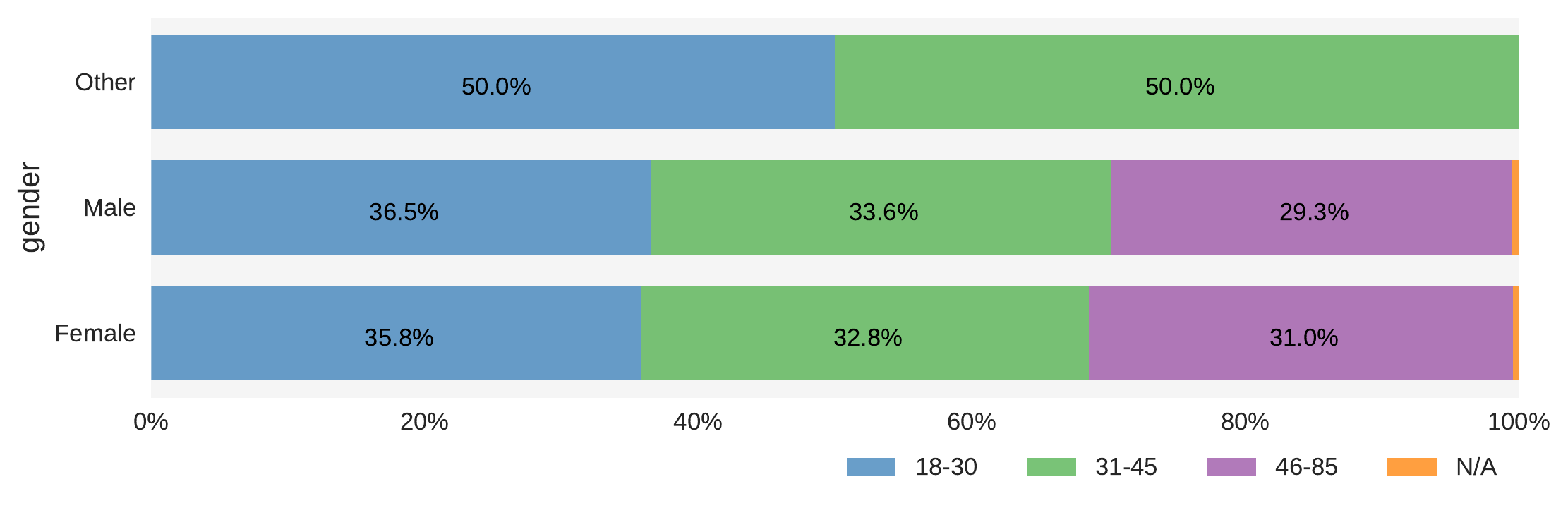}
        \caption{\small \textbf{Age} breakdown by \textbf{Gender}}\label{fig:age_gender}
    \end{subfigure}
    &
    \begin{subfigure}{\scale\textwidth}
        \includegraphics[width=\linewidth, height=3cm]{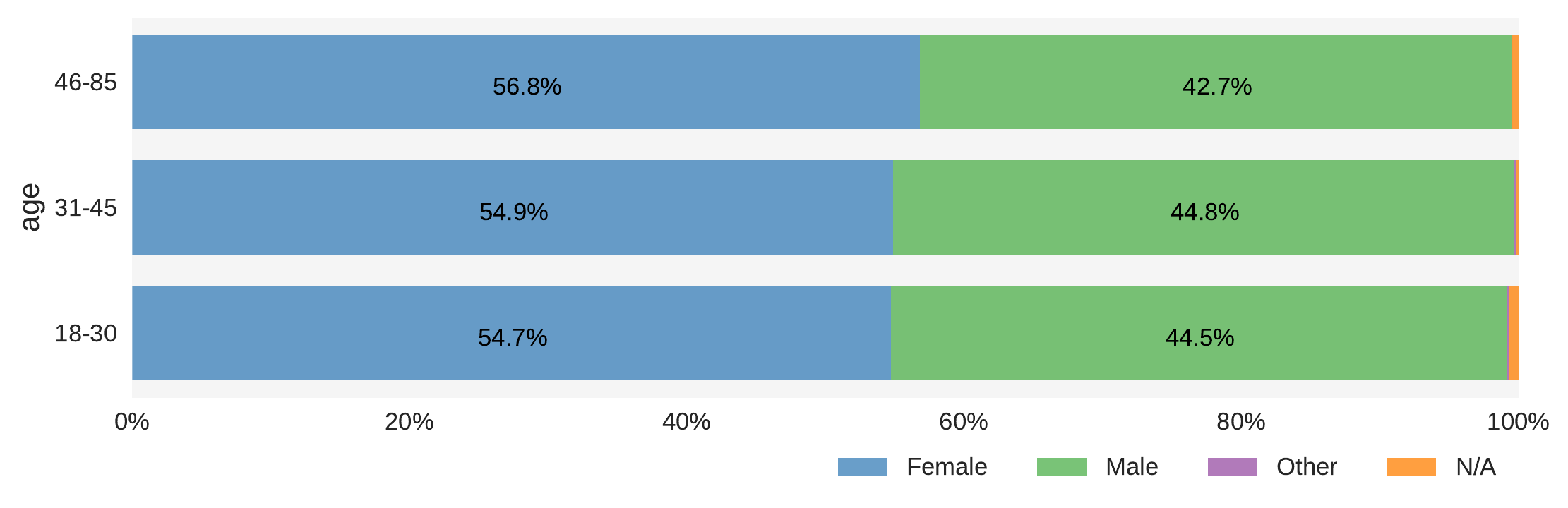}
        \caption{\small \textbf{Gender} breakdown by \textbf{Age}}\label{fig:gender_age}
    \end{subfigure}
    \\
    \begin{subfigure}{\scale\textwidth}
        \includegraphics[width=\linewidth, height=3cm]{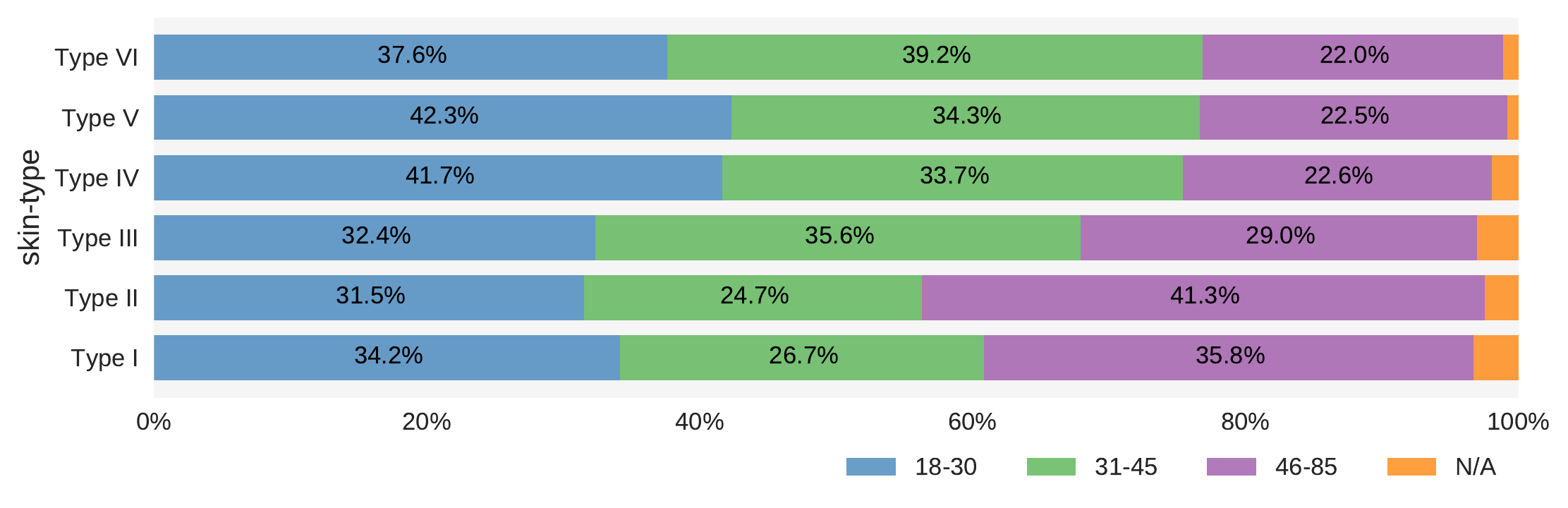}
        \caption{\small \textbf{Age} breakdown by \textbf{Skin type}}\label{fig:age_skin-type}
    \end{subfigure}
    &
    \begin{subfigure}{\scale\textwidth}
        \includegraphics[width=\linewidth, height=3cm]{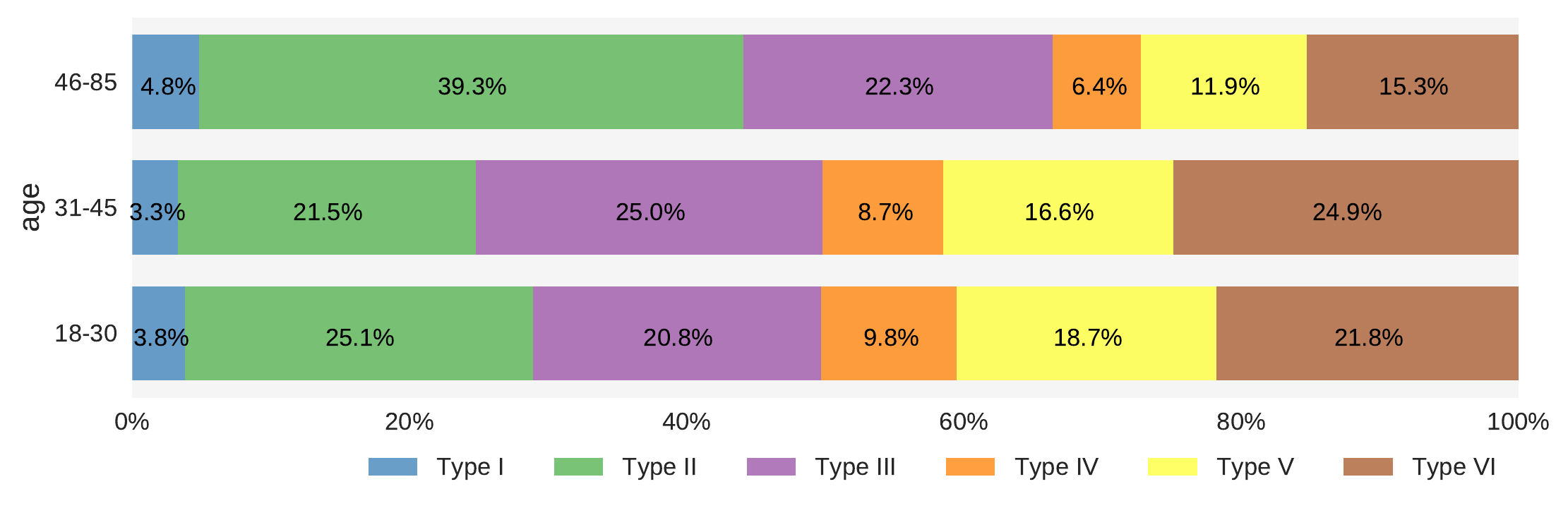}
        \caption{\small \textbf{Skin type} breakdown by \textbf{Age}}\label{fig:skin-type_age}
    \end{subfigure}
    \\
    \begin{subfigure}{\scale\textwidth}
        \includegraphics[width=\linewidth, height=3cm]{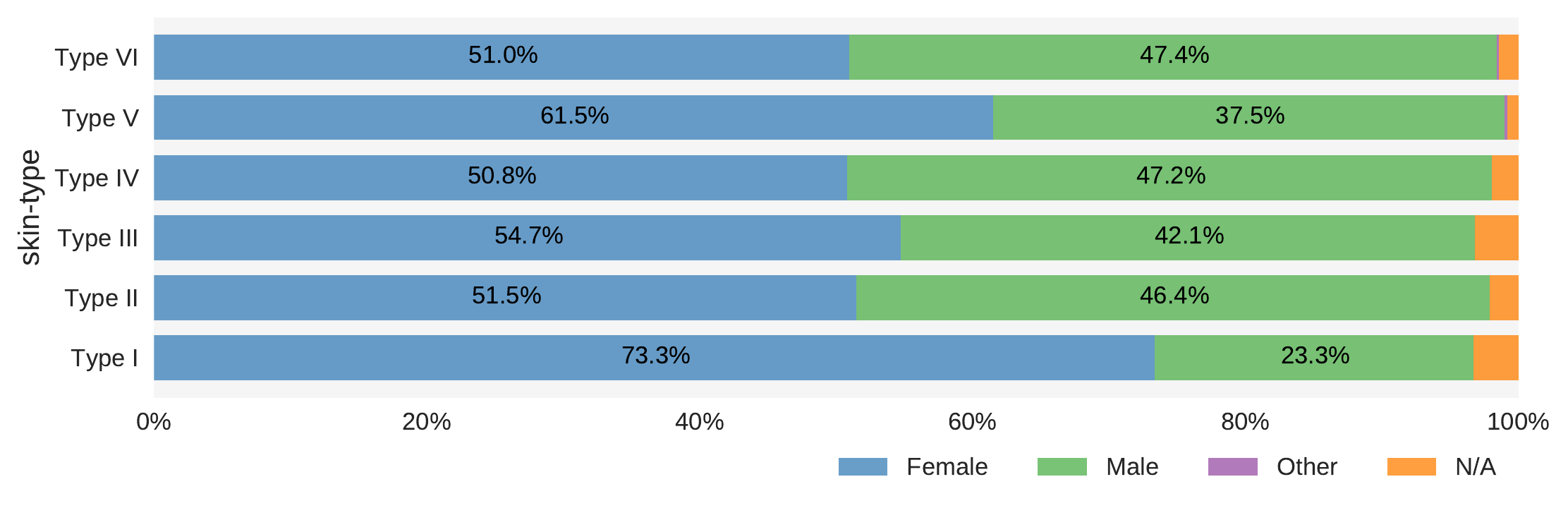}
        \caption{\small \textbf{Gender} breakdown by \textbf{Skin type}}\label{fig:gender_skin-type}
    \end{subfigure}
    &
    \begin{subfigure}{\scale\textwidth}
        \includegraphics[width=\linewidth, height=3cm]{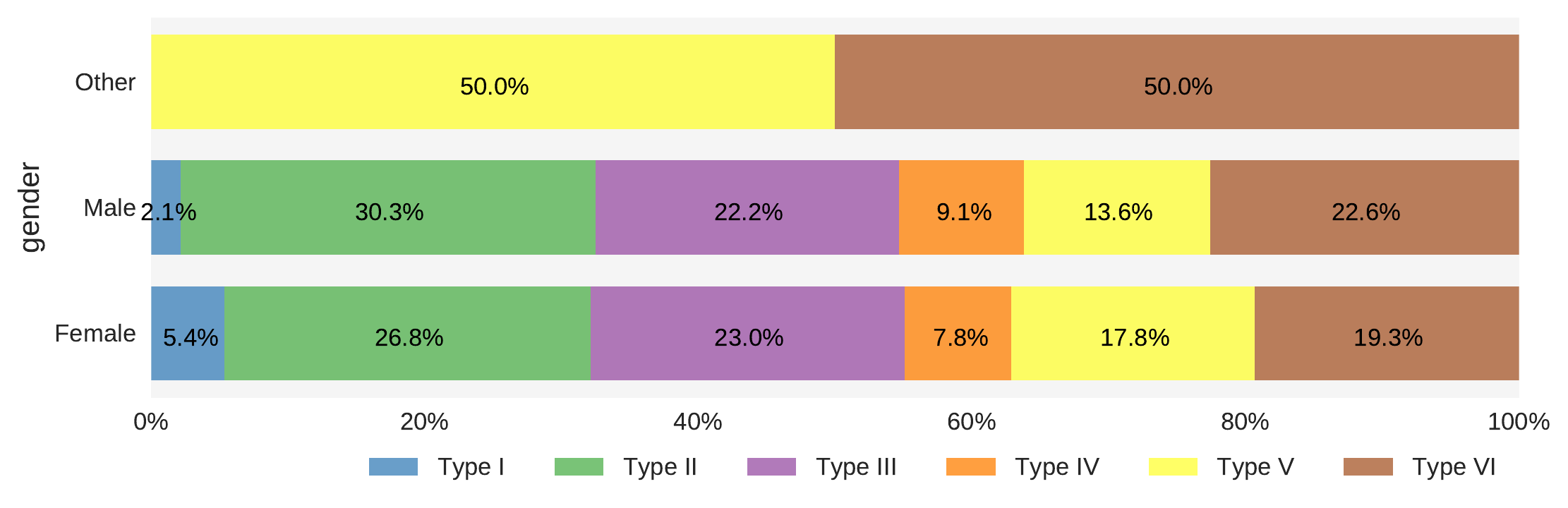}
        \caption{\small \textbf{Skin type} breakdown by \textbf{Gender}}\label{fig:skin-type_gender}
    \end{subfigure}
    \\
    \begin{subfigure}{\scale\textwidth}
        \includegraphics[width=\linewidth, height=3cm]{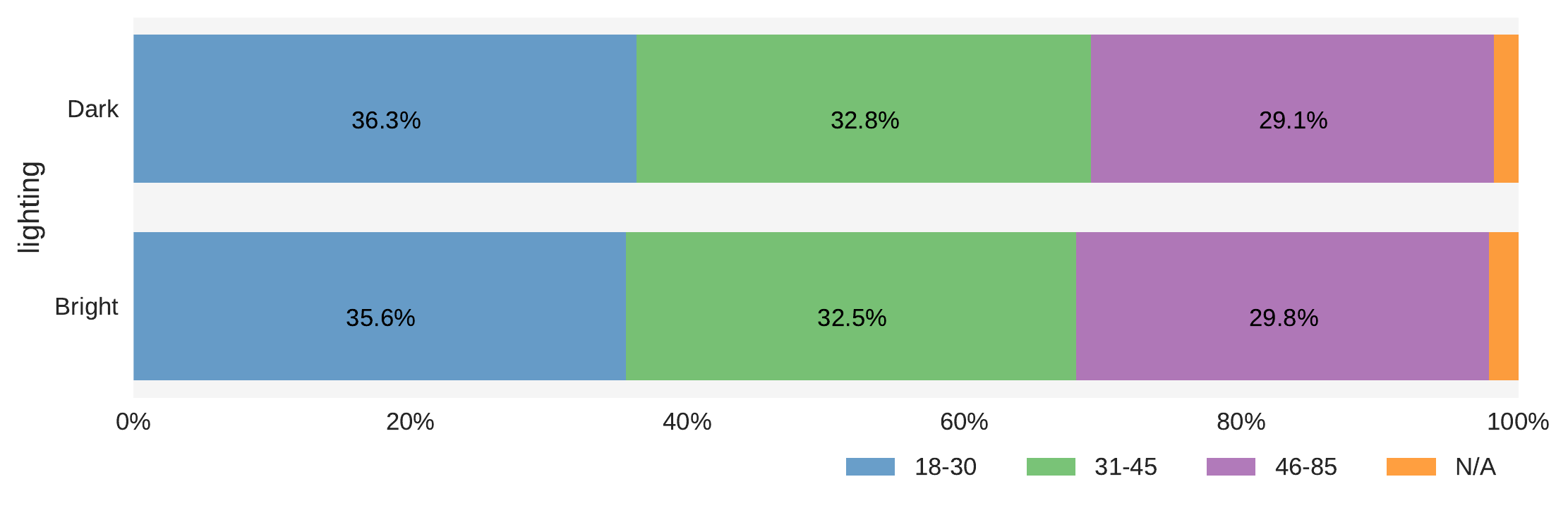}
        \caption{\small \textbf{Age} breakdown by \textbf{Lighting}}\label{fig:age_lighting}
    \end{subfigure}
    &
    \begin{subfigure}{\scale\textwidth}
        \includegraphics[width=\linewidth, height=3cm]{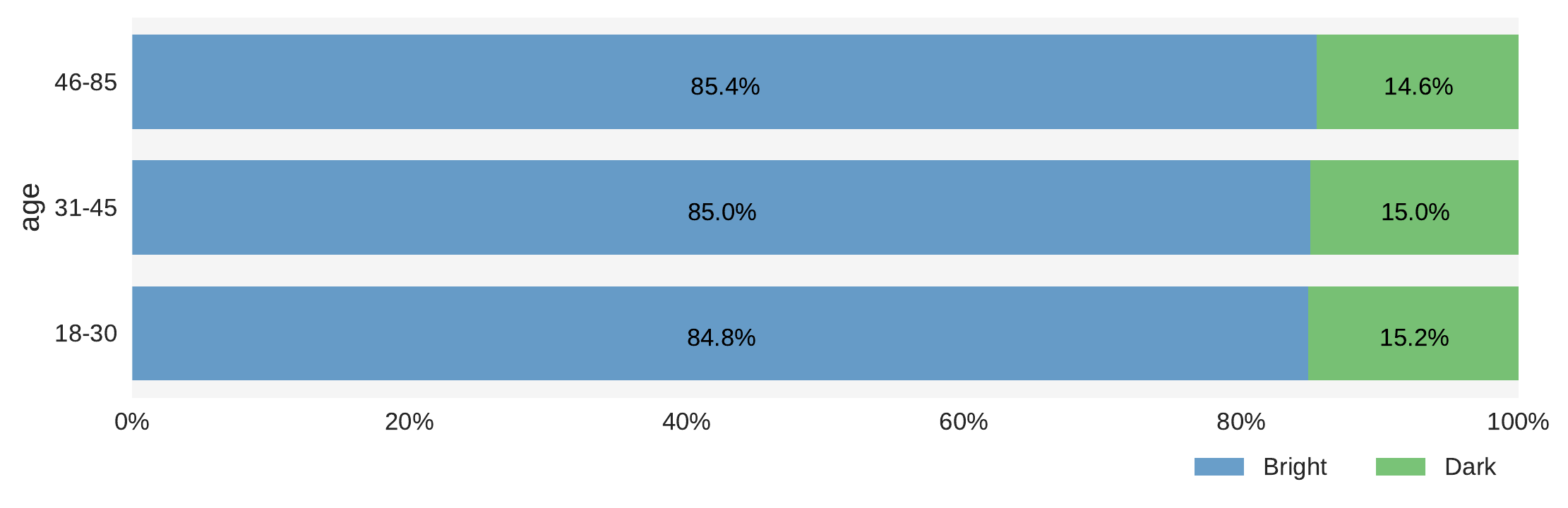}
        \caption{\small \textbf{Lighting} breakdown by \textbf{Age}}\label{fig:lighting_age}
    \end{subfigure}
    \\
    \begin{subfigure}{\scale\textwidth}
        \includegraphics[width=\linewidth, height=3cm]{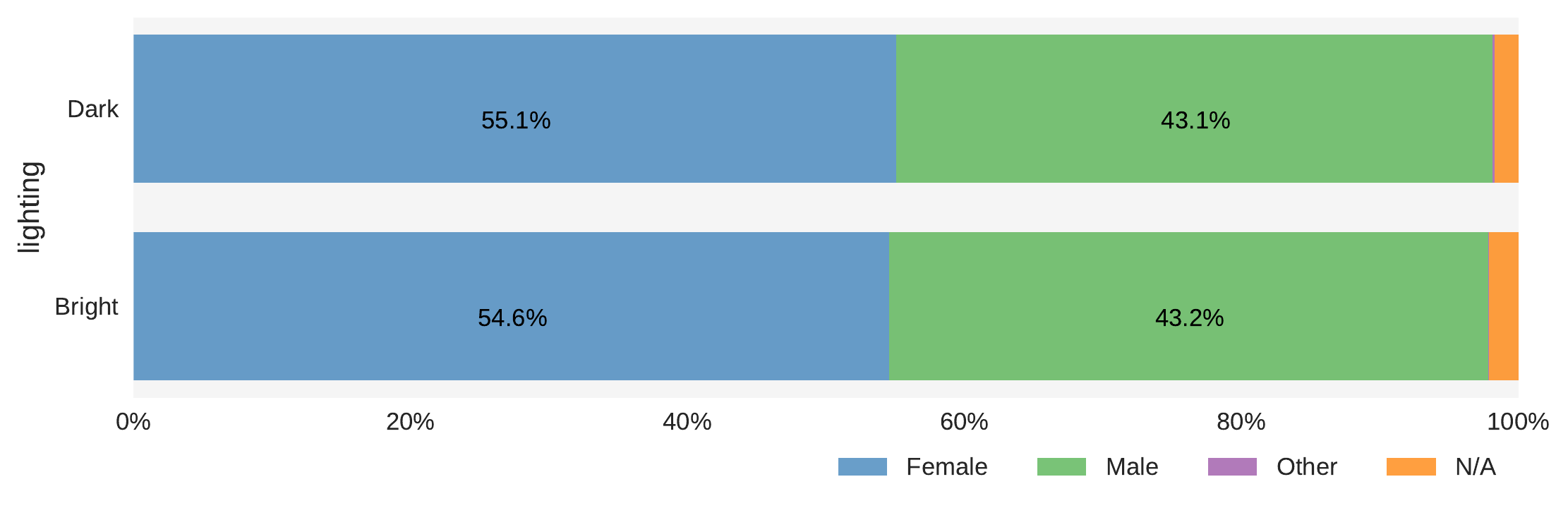}
        \caption{\small \textbf{Gender} breakdown by \textbf{Lighting}}\label{fig:gender_lighting}
    \end{subfigure}
    &
    \begin{subfigure}{\scale\textwidth}
        \includegraphics[width=\linewidth, height=3cm]{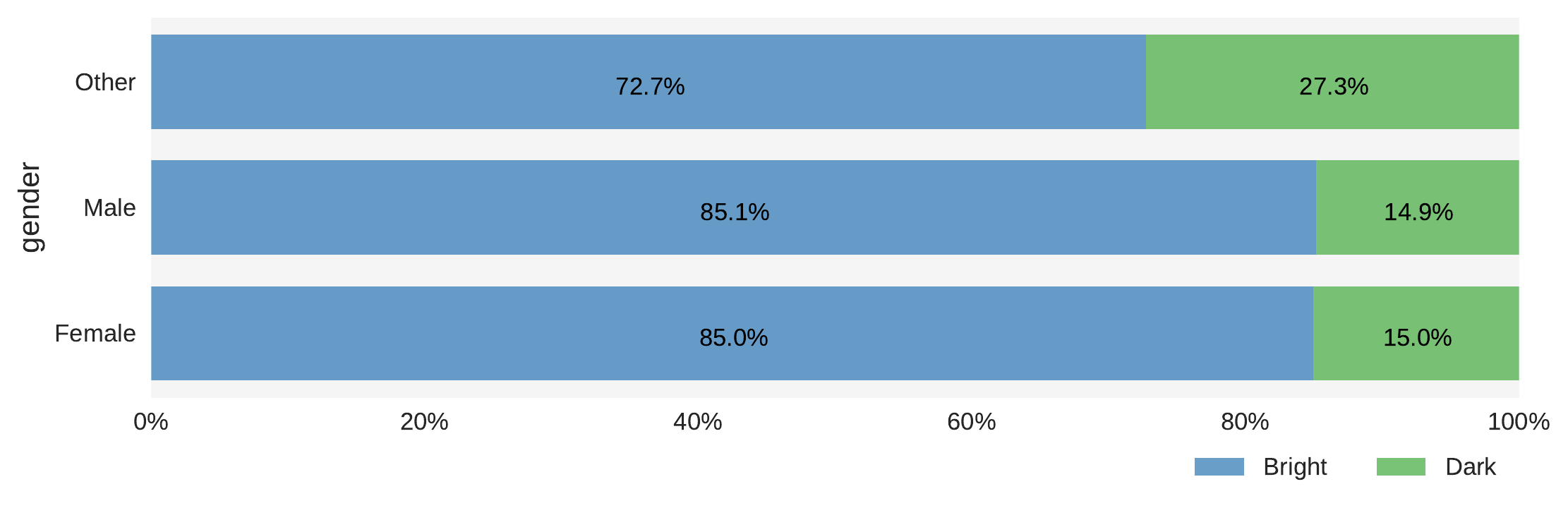}
        \caption{\small \textbf{Lighting} breakdown by \textbf{Gender}}\label{fig:lighting_gender}
    \end{subfigure}
    \\
    \begin{subfigure}{\scale\textwidth}
        \includegraphics[width=\linewidth, height=3cm]{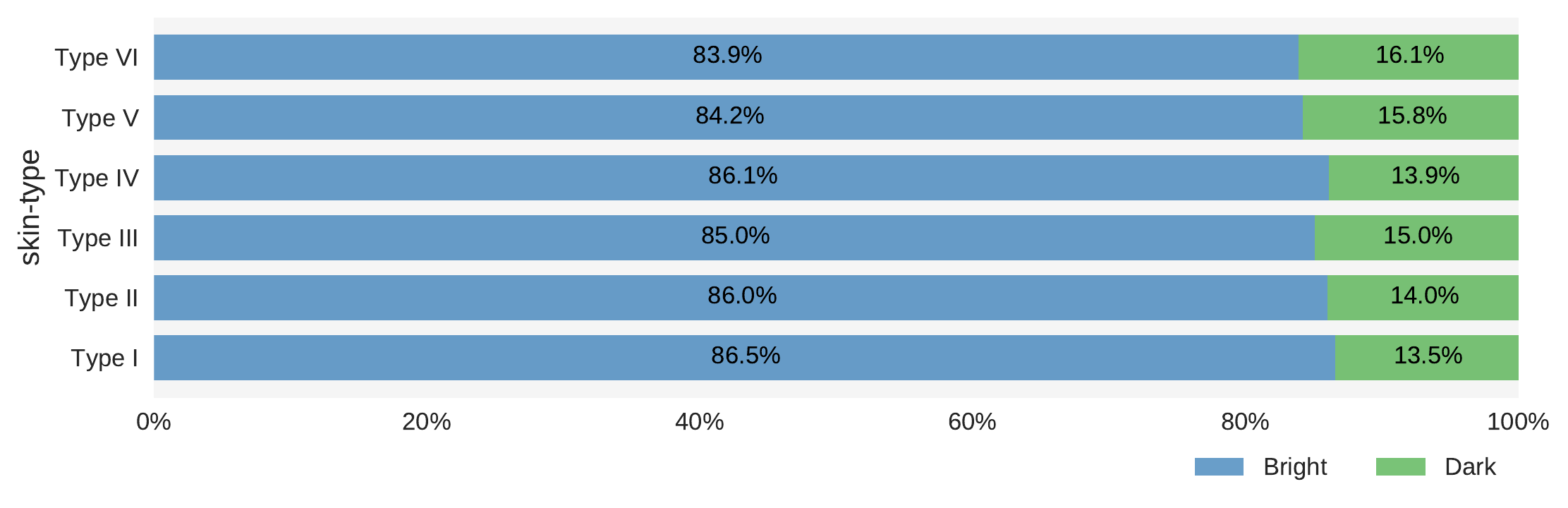}
        \caption{\small \textbf{Lighting} breakdown by \textbf{Skin type}}\label{fig:lighting_skin-type}
    \end{subfigure}
    &
    \begin{subfigure}{\scale\textwidth}
        \includegraphics[width=\linewidth, height=3cm]{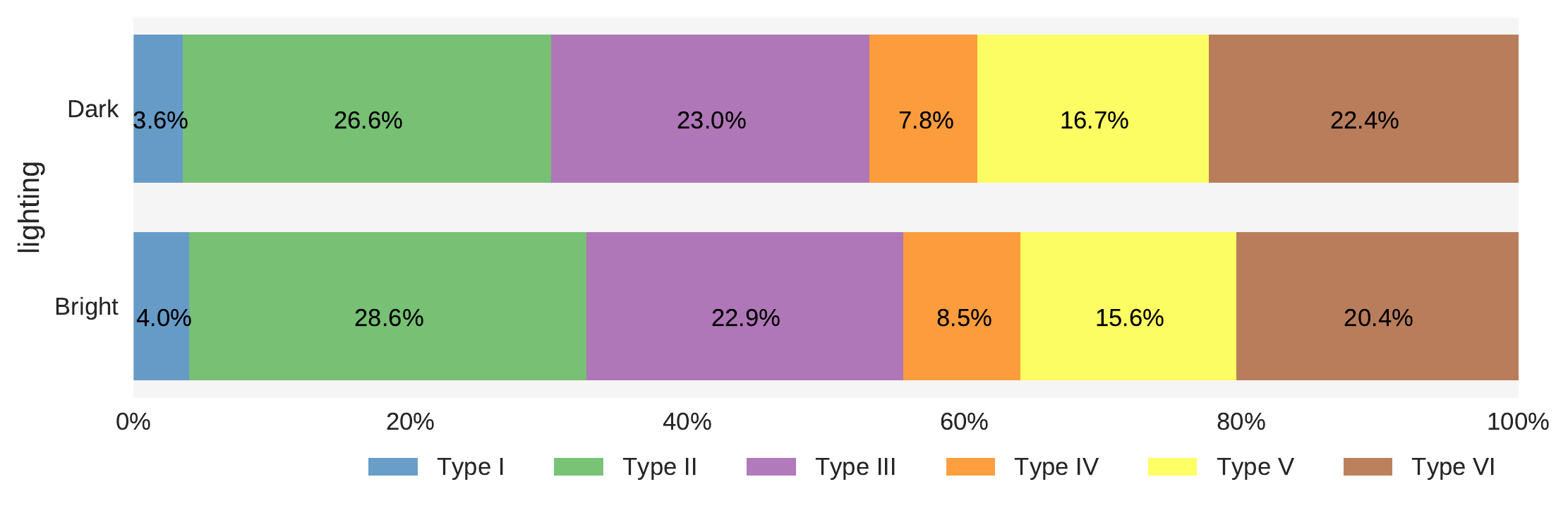}
        \caption{\small \textbf{Skin type} breakdown by \textbf{Lighting}}\label{fig:skin-type_lighting}
    \end{subfigure}
    \end{tabular}
\caption{\label{fig:grouped_dist}\textbf{Casual Conversations dataset} per-category distributions across subgroups defined by each of the remaining categories. The per-category distributions are generally well balanced across subgroups. }
\end{figure*}

\section{\label{sec:dataset}Casual Conversations Dataset}

The \textit{Casual Conversations} dataset is composed of approximately fifteen one-minute video recordings for each of our 3,011 subjects. Videos are captured in the United States in the cities of Atlanta, Houston, Miami, New Orleans, and Richmond. The subjects that participated in this study are from diverse age ($18+$), gender and skin tones. In most recordings, only one subject is present; however, there are videos in which two subjects are present simultaneously as well. Nonetheless, we only provide \textit{one set of labels} and it is for the current subject of interest.

In this dataset, we provide annotations for age, gender, apparent skin tone and whether or not the video was recorded in low ambient lighting. Age and gender attributes of subjects are provided by \textbf{\textit{subjects themselves}}. All other publicly available datasets~\textcolor{caner}{\cite{Eidinger14,Kaerkkaeinen19,Zhifei17}} provide hand or machine labelled \textcolor{caner}{or subjective~\cite{Buolamwini18}} annotations and therefore introduce a drastic bias towards appearance of a person other than the actual age and gender. Gender in our dataset is categorized as Male, Female, Other and N/A (preferred not to say or removed during data cleaning)~\textcolor{caner}{and refers to gender of subject at the time of collection}. We are aware that this categorization is over simplistic and does not sufficiently capture the diversity of genders that exist, and that we hope in the future there is more progress on enabling data analysis that captures this additional diversity while continuing to respect people’s privacy and any data ethics concerns.

In addition to \textbf{\textit{self-identified}} age and gender labels, we also provide skin tone annotations using the Fitzpatrick scale~\cite{FitzPatrick75}. Although the debate on ethnicity versus skin tone is still disputed ~\cite{Kaerkkaeinen19}, we believe it is \textcolor{caner}{subjective} considering that the apparent ethnicity of a person may differ from their actual ethnicity, thereby causing algorithms to classify incorrectly.~\textcolor{caner}{On the other hand, Hanna~\etal~\cite{Hanna2020} emphasizes that race is socially constructed and in algorithmic fairness methodologies  ML practitioners should be mindful of the dimensions of race that are relevant for particular contexts. In the context of computer vision, phenotypical representation is a particularly relevant dimension; we can acknowledge that measuring performance of computer vision models against socially constructed/self-identified categories might also be relevant and hope that other researchers are inspired to do so. Therefore, we believe} that skin tone is an expressive and generic way to group people, which is necessary to measure  the bias of the-state-of-the-art methods.~\textcolor{caner}{Even though, research on the usability of the Fitzpatrick scale~\cite{FitzPatrick75} is still on-going and recent study~\cite{Howard2021} recommends that this scale is unreliable and also a poor descriptor for skin tone, it has been} commonly used in classification of apparent skin tones. The Fitzpatrick scale constitutes six skin types based on the skin's reaction to Ultraviolet light. The scale ranges from Type I to VI, where Type I skin is pale to fair, never tans but always burns whereas Type VI skin is very dark, always tans but never burns (see example face crops in Figure~\ref{fig:visual1_hero_still}). Additionally, the Fitzpatrick scale has limitations in capturing diversity outside of the Western theories of race-related skin tone and does not perform as well for people with darker skin tones~\cite{Sambasivan20,Ware20,Howard2021}. Three of out the six skin types cover white skin, two cover brown skin, and there is only skin type for black skin, which clearly does not encompass the diversity within brown and black skin tones. A common procedure to alleviate this bias is to group the Fitzpatrick skin types into three buckets of light [types I, II], medium [types III \& IV], and dark skins [type V \& VI]. Our annotations provide the full, non-bucketed skin types such that others can decide how they'd to group the skin types.

In order to annotate for apparent skin types, eight individuals (raters) were appointed to annotate all subjects and to also flag the subjects that they are not confident about. As the final skin type annotations, we accumulated the weighted histograms over eight votes (uncertain votes are counted as half) and pick the most voted skin type as the ground-truth annotation.

Figure~\ref{fig:dist} shows the per-category distributions over our 3,011 subjects. As shown in the figures, we have decently balanced distributions over gender and age groups. For the skin type annotations, each paired group of types I \& II, III \& IV and V \& VI would be almost equal to one-third of the dataset. Uniform distributions of the annotations allow us to reduce the impact of bias in our measurements and hence let us better evaluate model robustness.

In Figure~\ref{fig:dist} the percentage of bright versus dark videos over all 45,186 videos is also depicted. To have a balanced lighting distribution, we sub-sample our dataset to include only one pair of videos per subject, a total of 6,022 videos. When possible, we chose one \textit{dark} and one \textit{bright} video. Note that sub-sampling only affects the lighting distribution because there is only one set of labels per subject in the dataset. After re-sampling, we end up with 37.3\% \textit{dark} videos in the smaller dataset. In all experiments, we use the mini Casual Conversations dataset except in DFDC evaluation (Section~\ref{sec:experiments}).

Although we have nearly uniform distributions per category, it is still very important to preserve uniformity in paired breakdowns. Figure~\ref{fig:grouped_dist} shows the paired distributions of categories. For example, in~Figures~\ref{fig:age_gender},~\ref{fig:gender_skin-type} and~\ref{fig:age_lighting}, all distributions are fairly uniform over all subcategories. In the rest of the paper, we refer to our four dimensional attributes as \textit{fairness} annotations.

Our dataset will be publicly available\footnote{\url{https://ai.facebook.com/datasets/casual-conversations-dataset}} for general use and we encourage users to extend annotations of our dataset for various computer vision applications, in line with our data use agreement.~\textcolor{caner}{We also would like to note that considering the limitations of our dataset,~\eg collected in one country and relatively smaller than the other datasets, we limit the use of it for only evaluation purposes. Therefore, our dataset \textbf{cannot} be used to train any model with the provided labels (see license agreement\footnote{\url{https://ai.facebook.com/datasets/casual-conversations-downloads}}).}

\section{\label{sec:experiments}Experiments}

The DeepFake Detection Challenge (DFDC)~\cite{DFDCPreview, DFDC2020} provided an opportunity for researchers to develop robust deepfake detection models and test on a challenging private test set. The top five winners of the competition had relatively low performance on the dataset, achieving only up to 70\% accuracy. However, in the scope of DFDC, AI models were only evaluated on a binary classification task,~\ie whether a video is fake or not. Since a portion of the DFDC private test set is constructed using videos from Casual Conversations, to complete the missing dimension of DFDC, we match the overlapping 4,945 DFDC test videos (almost half of the private test set) with their ground truth fairness annotations (age, gender, apparent skin tone and lighting) and display the ROC curves of the top five winners on each dimension in Figure~\ref{fig:dfdc_roc_curves}. Interestingly, we found that in terms of balanced predictions the third placed winner, NTechLab~\cite{NTechLab2020}, performs more consistently across three dimensions (age, gender, lighting) in comparison to all other winners.

\begin{table*}[ht!]
    \centering
    \resizebox{\textwidth}{!}{ 
    \arrayrulecolor{lightgray}
    \begin{tabular}{lc|*{3}{c}|*{2}{c}|*{6}{c}|*{2}{c}}
        \toprule
        \multicolumn{1}{c}{} & \multicolumn{1}{c}{} & \multicolumn{3}{c}{\textbf{Age}} & \multicolumn{2}{c}{\textbf{Gender}} & \multicolumn{6}{c}{\textbf{Skin type}} & \multicolumn{2}{c}{\textbf{Lighting}} \\
        \toprule
        Winner & Overall & 18-30 & 31-45 & 46-85 & Female & Male & Type I & Type II & Type III & Type IV & Type V & Type VI & Bright & Dark \\
        \midrule
        Selim Seferbekov~\cite{Seferbekov2020} & -2.500 & -2.437 & -2.476 & -2.610 & -2.443 & -2.569 & -1.427 & -2.356 & -2.360 & -1.851 & -2.714 & -3.098 & -2.569 & -2.184 \\
        WM~\cite{WM2020} & -2.276 & -2.140 & -2.316 & -2.337 & -2.319 & -2.140 & -0.814 & -2.113 & -2.303 & -1.792 & -2.567 & -2.649 & -2.362 & -1.865 \\
        NTechLab~\cite{NTechLab2020} & -2.008 & -2.008 & -2.007 & -2.023 & -2.050 & -1.914 & -1.417 & -1.729 & -2.047 & -1.357 & -2.253 & -2.497 & -2.081 & -1.668 \\
        Eighteen Years Old~\cite{Eighteen2020} & -2.027 & -2.112 & -2.110 & -1.876 & -2.026 & -1.996 & -1.081 & -1.764 & -1.975 & -1.095 & -2.288 & -2.651 & -2.108 & -1.647 \\
        The Medics~\cite{Medics2020} & -2.688 & -2.453 & -2.604 & -2.877 & -2.597 & -2.719 & -1.677 & -2.598 & -2.678 & -2.303 & -2.790 & -3.115 & -2.732 & -2.501 \\
        \bottomrule
    \end{tabular}}
    \caption{\textbf{Precision} comparison of the DFDC winners, with a breakdown by fairness categories. Log of the weighted precision~\cite{DFDC2020} is reported. Note that overall scores slightly differ from those reported in~\cite{DFDC2020} due to the number of examples (see Section~\ref{sec:experiments}).}
    \label{tab:dfdc_precision}
\end{table*}

In Table~\ref{tab:dfdc_precision}, we present the log of the weighted precision~\cite{DFDC2020} per category. Surprisingly, the best performing method on the subset of the private test set is NTechLab~\cite{NTechLab2020} as opposed to the top winner Selim Seferbekov~\cite{Seferbekov2020}. However, despite higher performance, the results show that all models except NTechLab's~\cite{NTechLab2020} are particularly biased towards paler skin tones (\textit{Type~I}). Nevertheless, all winning approaches struggle to identify fake videos containing subjects with darker skin tones (\textit{Type V} and \textit{VI}). Aside from skin tone, the data indicate that Eighteen Years Old~\cite{Eighteen2020} significantly outperforms other winning models on older individuals (age $\in [46, 85]$).

\begin{table*}[ht!]
    \centering
    \resizebox{\textwidth}{!}{ 
    \arrayrulecolor{lightgray}
    \begin{tabular}{lc|*{3}{c}|*{2}{c}|*{6}{c}|*{2}{c}}
        \toprule
        \multicolumn{1}{c}{} & \multicolumn{1}{c}{} & \multicolumn{3}{c}{\textbf{Age}} & \multicolumn{2}{c}{\textbf{Gender}} & \multicolumn{6}{c}{\textbf{Skin type}} & \multicolumn{2}{c}{\textbf{Lighting}} \\
        \toprule
        Winner & Overall & 18-30 & 31-45 & 46-85 & Female & Male & Type I & Type II & Type III & Type IV & Type V & Type VI & Bright & Dark \\
        \midrule
        Selim Seferbekov~\cite{Seferbekov2020} & 0.195 & 0.201 & 0.184 & 0.203 & 0.185 & 0.210 & 0.147 & 0.180 & 0.173 & 0.148 & 0.208 & 0.265 & 0.198 & 0.178 \\
        WM~\cite{WM2020} & 0.176 & 0.174 & 0.172 & 0.184 & 0.185 & 0.163 & 0.110 & 0.176 & 0.175 & 0.146 & 0.183 & 0.202 & 0.179 & 0.163 \\
        NTechLab~\cite{NTechLab2020} & 0.166 & 0.175 & 0.169 & 0.159 & 0.168 & 0.166 & 0.142 & 0.154 & 0.165 & 0.099 & 0.176 & 0.209 & 0.164 & 0.171 \\
        Eighteen Years Old~\cite{Eighteen2020} & 0.186 & 0.205 & 0.180 & 0.178 & 0.193 & 0.175 & 0.143 & 0.179 & 0.204 & 0.144 & 0.177 & 0.209 & 0.187 & 0.183 \\
        The Medics~\cite{Medics2020} & 0.213 & 0.204 & 0.207 & 0.221 & 0.205 & 0.216 & 0.164 & 0.218 & 0.226 & 0.189 & 0.202 & 0.219 & 0.214 & 0.211 \\
        \bottomrule
            \end{tabular}}
    \caption{\textbf{Log-loss} comparison of the DFDC winners, with a breakdown by fairness categories. These results showcase that it is more difficult for all methods to identify deepfakes when darker skin tones are present.}
    \label{tab:dfdc_logloss}
\end{table*}

\begin{figure}
    \resizebox{\linewidth}{!}{
        \begin{tabular}{*{5}c}
            \includegraphics{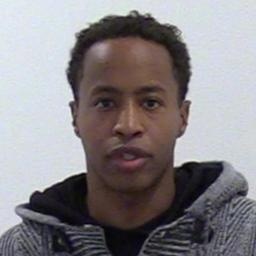} &
            \includegraphics{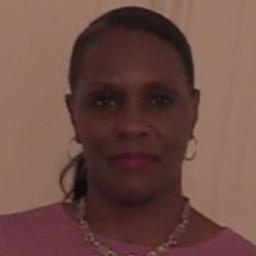} &
            \includegraphics{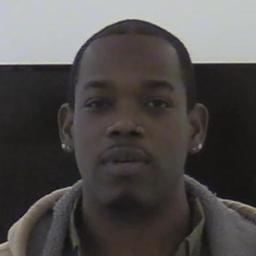} &
            \includegraphics{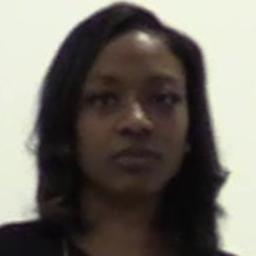} &
            \includegraphics{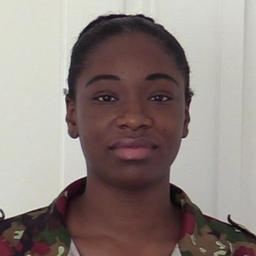}
            \\
            \includegraphics{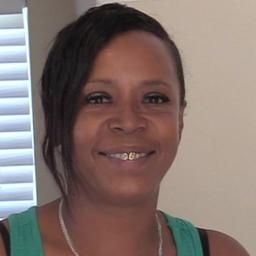} &
            \includegraphics{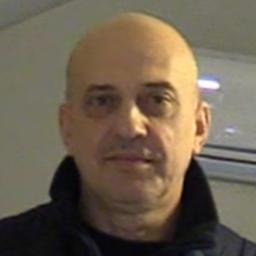} &
            \includegraphics{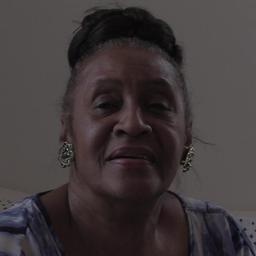} &
            \includegraphics{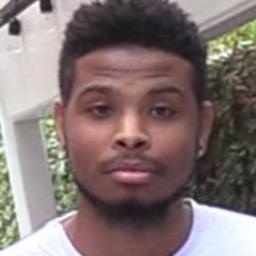} &
            \includegraphics{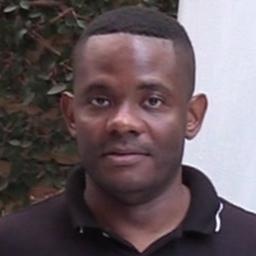}
            \\
            \includegraphics{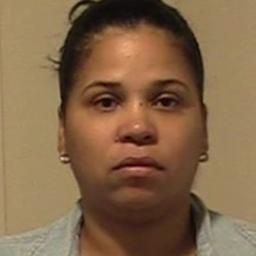} &
            \includegraphics{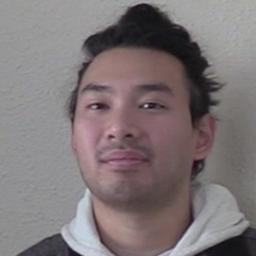} &
            \includegraphics{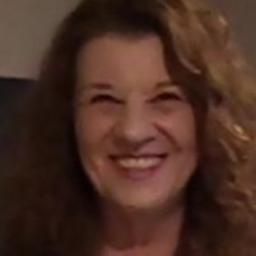} &
            \includegraphics{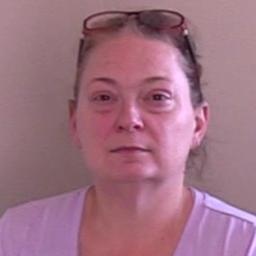} &
            \includegraphics{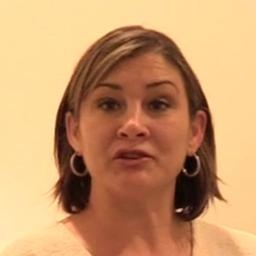}
        \end{tabular}
    }
    \caption{\label{fig:dfdc_fn}\textbf{Example face crops} of False Negatives (FNs), where all methods labelled the videos as \textbf{fake} although they are benign examples. Methods mostly fail on the darker skin-tones. \textcolor{caner}{Note that we show the \textit{unperturbed} (benign) face crops in the figure.}}
\end{figure}

\begin{figure}
    \resizebox{\linewidth}{!}{
        
        \begin{tabular}{*{5}c}  
            \includegraphics{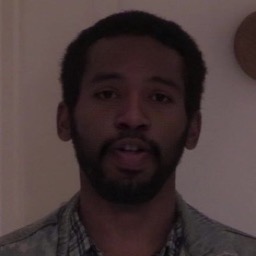} & 
            \includegraphics{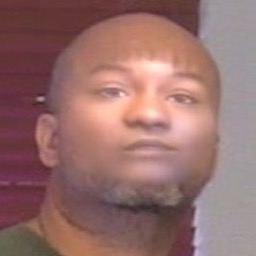} & 
            \includegraphics{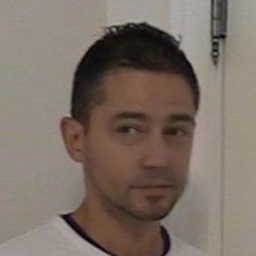} & 
            \includegraphics{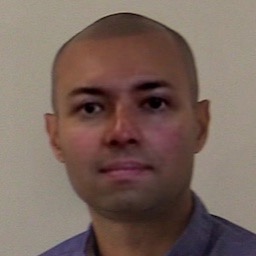} & 
            \includegraphics{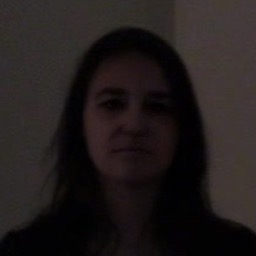}
            \\
            \includegraphics{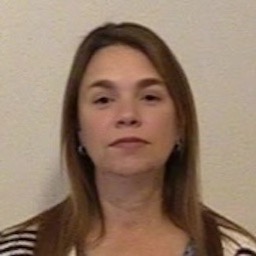} & 
            \includegraphics{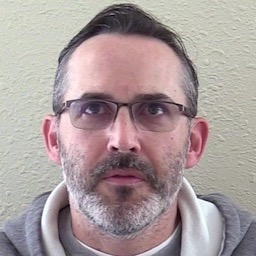} & 
            \includegraphics{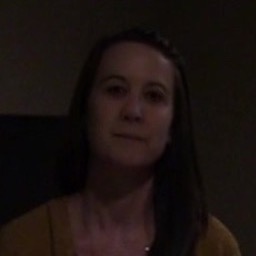} & 
            \includegraphics{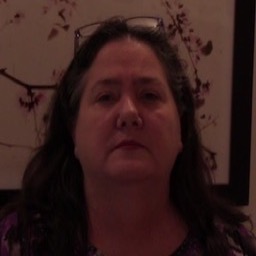} & 
            \includegraphics{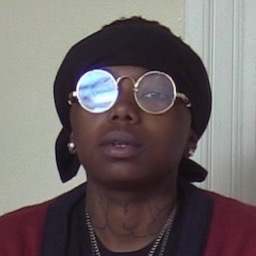}
        \end{tabular}
    }
    \caption{\textbf{Example face crops} of False Positives (FPs), where all methods labelled the videos as \textbf{real} although they are perturbed. Note that we show the \textcolor{caner}{\textit{perturbed} (fake)} face crops in the figure.}
    \label{fig:dfdc_fp}
\end{figure}

In addition to precision, we also report the log-loss of each model per fairness category in Table~\ref{tab:dfdc_logloss}. Similar to the weighted precision results, NTechLab~\cite{NTechLab2020} and Eighteen Years Old~\cite{Eighteen2020} have the lowest overall log-loss. As expected, the log-loss of each methods increases dramatically on darker skin tones.

Lastly, we also investigated the False Negatives (FNs) of each method because video clips of those examples in the DFDC are benign \textit{real} samples without any face perturbations (may have some image augmentations applied). Therefore, we show the FN ratios of each winner per category in Table~\ref{tab:dfdc_fn}. We also present the FN ratios on the videos where all methods failed in the last row (intersection). Considering the distributions in Figure~\ref{fig:dist}, we conclude that most of the scores are in alignment with the data distribution except the darker skin tones,~\ie types V \& VI.

Figures~\ref{fig:dfdc_fn} and~\ref{fig:dfdc_fp} present face crops from the original video clips where all top winners failed to classify correctly. Figure~\ref{fig:dfdc_fn} shows example face crops for FNs. As also stated above, all top five winners struggle to identify the originality of videos when the subject in interest has a darker skin tone.

\begin{table*}[ht!]
    \centering
    \resizebox{\textwidth}{!}{ 
    \arrayrulecolor{lightgray}
    \begin{tabular}{lc|*{3}{c}|*{2}{c}|*{6}{c}|*{2}{c}}
        \toprule
        \multicolumn{1}{c}{} & \multicolumn{1}{c}{} & \multicolumn{3}{c}{\textbf{Age}} & \multicolumn{2}{c}{\textbf{Gender}} & \multicolumn{6}{c}{\textbf{Skin type}} & \multicolumn{2}{c}{\textbf{Lighting}} \\
        \toprule
        Winner & \# & 18-30 & 31-45 & 46-85 & Female & Male & Type I & Type II & Type III & Type IV & Type V & Type VI & Bright & Dark \\
        \midrule
         Selim Seferbekov~\cite{Seferbekov2020} & 246 & 28.46 & 33.33 & 36.99 & 54.07 & 43.90 & 2.03 & 23.58 & 22.76 & 4.07 & 20.33 & 27.24 & 84.55 & 15.45 \\
         WM~\cite{WM2020}                       & 190 & 26.32 & 36.84 & 33.68 & 60.00 & 35.26 & 1.05 & 23.68 & 27.37 & 4.74 & 21.05 & 22.11 & 86.84 & 13.16 \\
         NTechLab~\cite{NTechLab2020}           & 142 & 30.99 & 32.39 & 34.51 & 59.15 & 37.32 & 3.52 & 19.72 & 28.17 & 4.23 & 19.72 & 24.65 & 85.92 & 14.08 \\
         Eighteen Years Old~\cite{Eighteen2020} & 141 & 31.91 & 39.01 & 27.66 & 56.74 & 40.43 & 2.13 & 19.86 & 24.82 & 2.84 & 21.99 & 28.37 & 86.52 & 13.48 \\
         The Medics~\cite{Medics2020}           & 307 & 23.13 & 31.92 & 40.39 & 53.75 & 41.04 & 2.28 & 24.43 & 25.41 & 5.86 & 18.24 & 23.78 & 82.08 & 17.92 \\
         intersection                           &  38 & 23.68 & 44.74 & 28.95 & 55.26 & 39.47 & 5.26 & 13.16 & 28.95 & 2.63 & 21.05 & 28.95 & 84.21 & 15.79 \\
        \bottomrule
            \end{tabular}}
    \caption{\textbf{False Negative (FN) ratios} of the DFDC winners, with a breakdown by fairness categories. Scores in category are normalized by the number of examples. The data above shows that methods performs significantly worse on examples of darker-skinned subjects, considering the skin type distribution in Figure~\ref{fig:dist}.}
    \label{tab:dfdc_fn}
\end{table*}

\textcolor{caner}{There are several reasons why methods are not performing equally well for each subgroups. The~\textit{Casual Conversations dataset} is only one portion of the DFDC and the other portion may have larger imbalance~\wrt age, gender and skin tone. Another performance drop might be induced at training time due to optimization schemes, such as how images were sampled, if augmentations were applied on the input. In addition, our annotations were not provided in DFDC and models trained without fairness labels and considering fairness in mind, would naturally perform worse on groups of people where there is significantly less number of samples.} 

\subsection{Apparent age and gender prediction}
We compared the apparent age and gender prediction results of the three state-of-the-art models, evaluated on our dataset. In the following experiments, we used the reduced (mini) dataset. We first detect faces in each frame with DLIB~\cite{King2009} and evaluate the models on the sampled 100 face crops per video. Final predictions are calculated by aggregating results over these samples (most voted gender and median age). Levi \& Hassner~\cite{LeviHassner15} and LMTCNN~\cite{Lee2018} predicts age in predefined brackets and therefore we map their age prediction to our predefined age groups in Figure~\ref{fig:dist}.

Tables~\ref{tab:age_precision} and~\ref{tab:gender_precision} show the precision of the models on apparent age and gender, respectively. Levi \& Hassner~\cite{LeviHassner15} is one of the early works that used deep neural networks. It is comparatively less accurate method among all, however, almost as good in apparent gender classification as LMTCNN~\cite{Lee2018}. LightFace~\cite{Serengil2020}, on the other hand, is more successful on predicting accurate apparent age and gender. Nevertheless, state-of-the-art methods' apparent gender precision on darker skin types (Type V \& VI) is drastically lower by more than 20\% on average.

\begin{table*}
    \resizebox{\textwidth}{!}{ 
    \arrayrulecolor{lightgray}
    \begin{tabular}{lc|*{3}{c}|*{6}{c}|*{2}{c}}
        \toprule
        \multicolumn{1}{c}{} & \multicolumn{1}{c}{} & \multicolumn{3}{c}{\textbf{Gender}} & \multicolumn{6}{c}{\textbf{Skin type}} & \multicolumn{2}{c}{\textbf{Lighting}} \\
        \toprule
        & Overall & Female & Male & Other & Type I & Type II & Type III & Type IV & Type V & Type VI & Bright & Dark \\
        \midrule
        Levi \& Hassner~\cite{LeviHassner15} & 38.05 & 37.44 & 39.48 & 66.67 & 39.56 & 38.72 & 40.84 & 36.47 & 36.47 & 34.89 & 38.49 & 37.04 \\
        LMTCNN~\cite{Lee2018} & 42.26 & 42.28 &  44.53 & 100.00 & 42.33 & 41.78 & 42.30 &   42.79 &  42.44 &   37.99 &  42.94 &  41.12 \\
        LightFace~\cite{Serengil2020} & 54.32 & 54.21 & 56.18 & 83.33 & 46.51 & 55.52 & 54.59 & 55.78 & 53.78 & 52.57 & 54.17 & 55.20 \\
        \bottomrule
            \end{tabular}}
    \caption{\textbf{Precision} comparison of the apparent \textbf{age classification} methods, with a breakdown by fairness categories.}
    \label{tab:age_precision}
\end{table*}

\begin{table*}
    \resizebox{\textwidth}{!}{ 
    \arrayrulecolor{lightgray}
    \begin{tabular}{lc|*{3}{c}|*{6}{c}|*{2}{c}}
        \toprule
        \multicolumn{1}{c}{} & \multicolumn{1}{c}{} & \multicolumn{3}{c}{\textbf{Age}} & \multicolumn{6}{c}{\textbf{Skin type}} & \multicolumn{2}{c}{\textbf{Lighting}} \\
        \toprule
        & Overall & 18-30 & 31-45 & 46-85 & Type I & Type II & Type III & Type IV & Type V & Type VI & Bright & Dark \\
        \midrule
        Levi \& Hassner~\cite{LeviHassner15} & 39.42 & 39.29 & 40.65 & 54.00 & 47.51 & 56.81 & 55.97 & 53.97 & 35.89 & 35.30 & 40.21 & 38.12 \\
        LMTCNN~\cite{Lee2018} & 41.56 & 41.57 & 43.14 & 56.29 & 50.38 & 58.64 & 58.22 & 55.85 & 38.62 & 39.68 & 41.44 & 41.75 \\
        LightFace~\cite{Serengil2020} & 44.12 & 44.33 & 45.40 & 59.85 & 55.73 & 62.39 & 61.12 & 61.81 & 41.90 & 41.66 & 44.29 & 43.86 \\
        \bottomrule
            \end{tabular}}
    \caption{\textbf{Precision} comparison of the apparent \textbf{gender classification} methods, with a breakdown by fairness categories.}
    \label{tab:gender_precision}
\end{table*}

\newcommand{\scaleroc}{0.22}
\begin{figure*}
    \begin{tabular}{m{0.1pt}ccccc}
    & Age & Gender & Skin type & Lighting
    \\
    \arrayrulecolor{lightgray}\cmidrule{2-5}
    \rotatebox[origin=l]{90}{\small Selim Seferbekov~\cite{Seferbekov2020}}
    &
    \begin{subfigure}{\scaleroc\linewidth}
        \includegraphics[width=\textwidth]{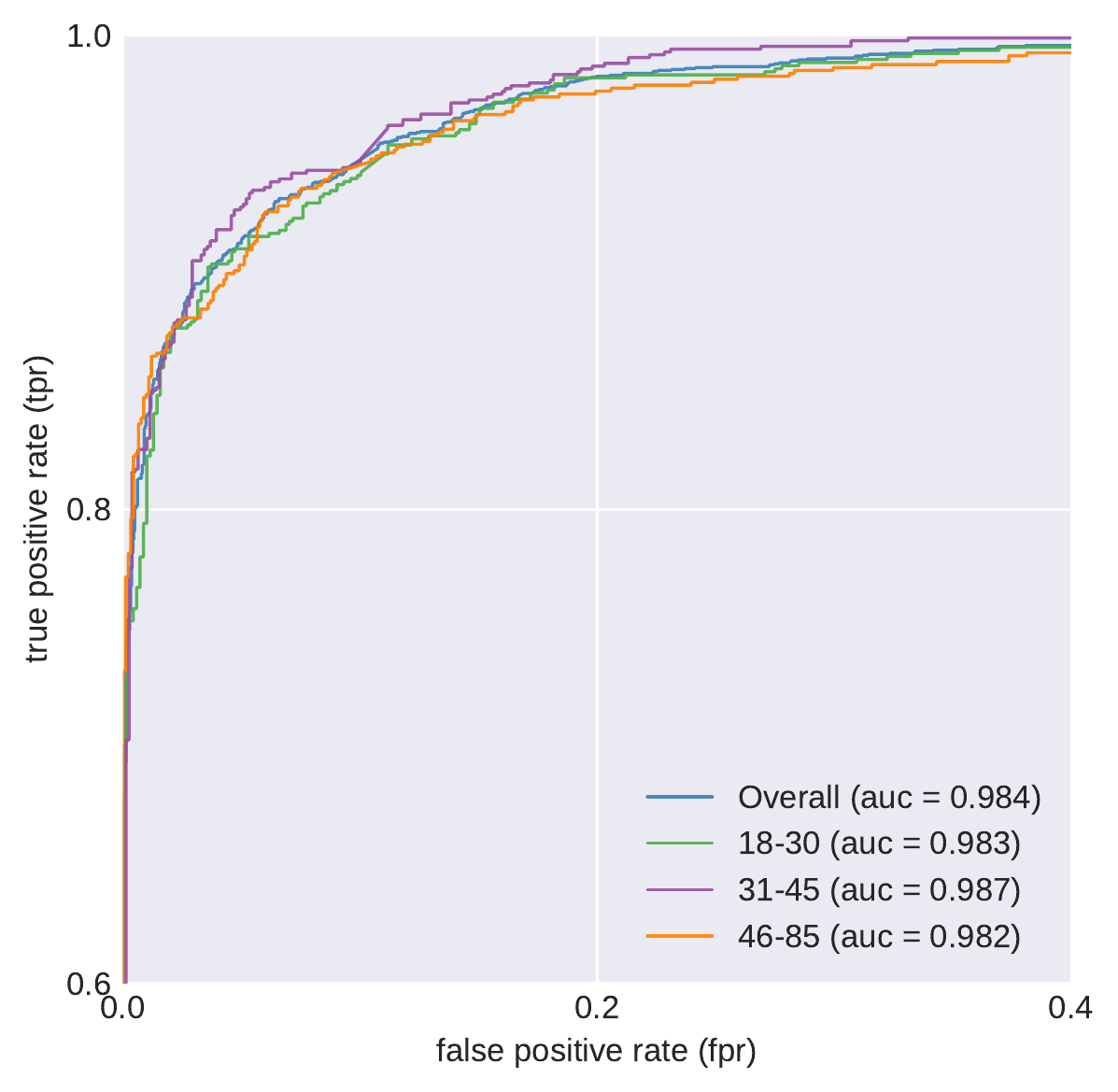}
    \end{subfigure}
    &
    \begin{subfigure}{\scaleroc\linewidth}
        \includegraphics[width=\textwidth]{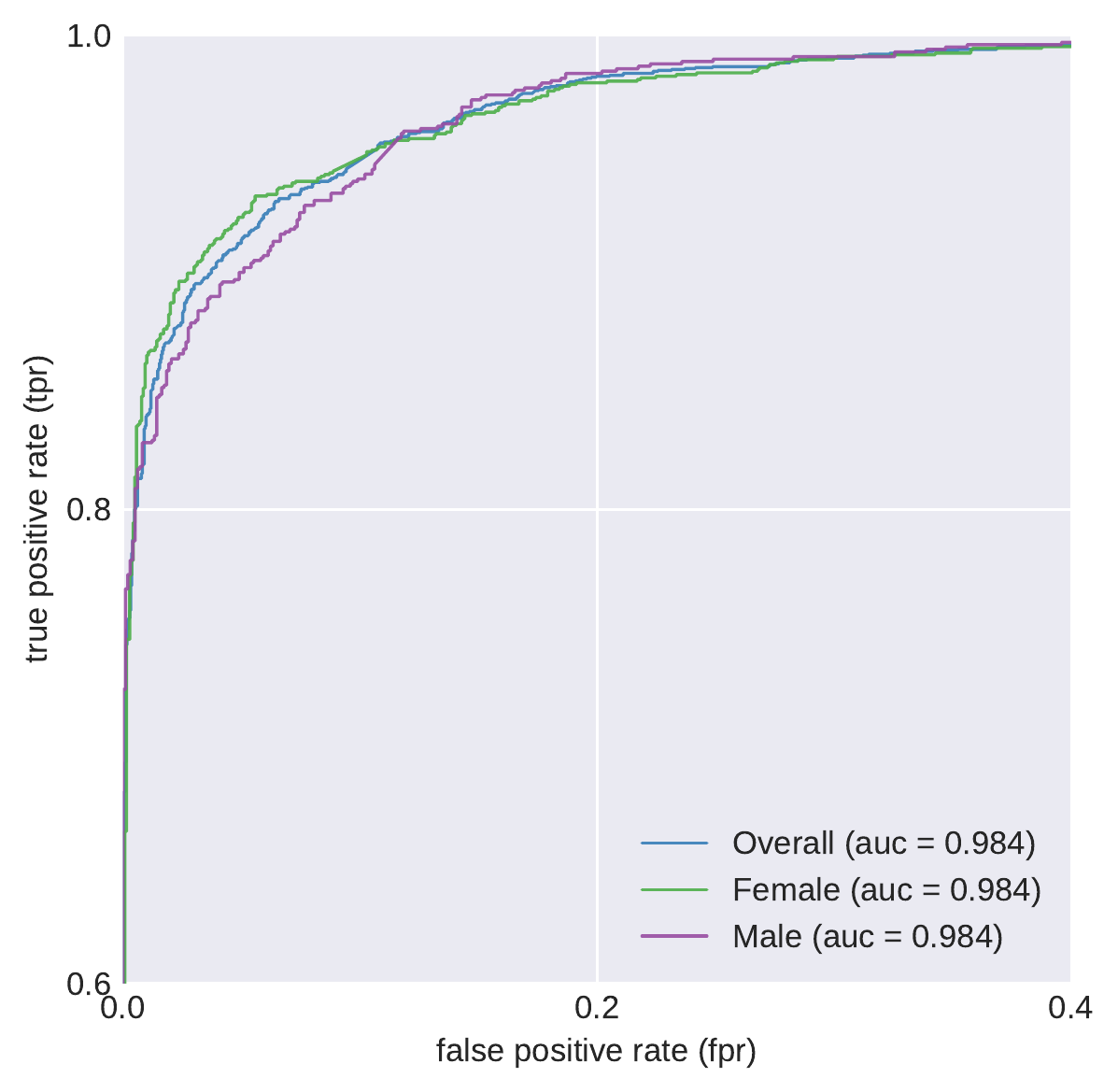}
    \end{subfigure}
    &
    \begin{subfigure}{\scaleroc\linewidth}
        \includegraphics[width=\textwidth]{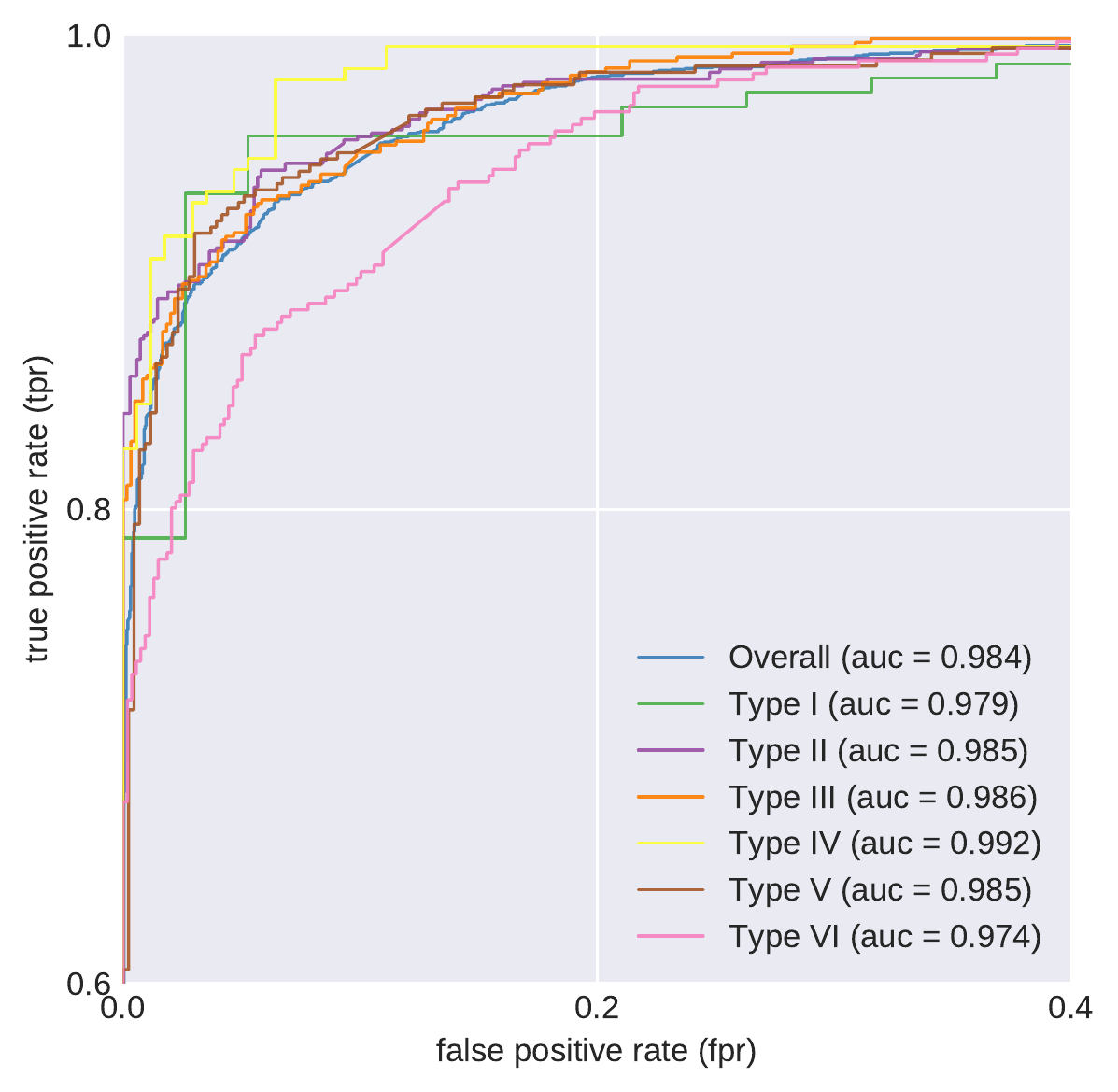}
    \end{subfigure}
    &
    \begin{subfigure}{\scaleroc\linewidth}
        \includegraphics[width=\textwidth]{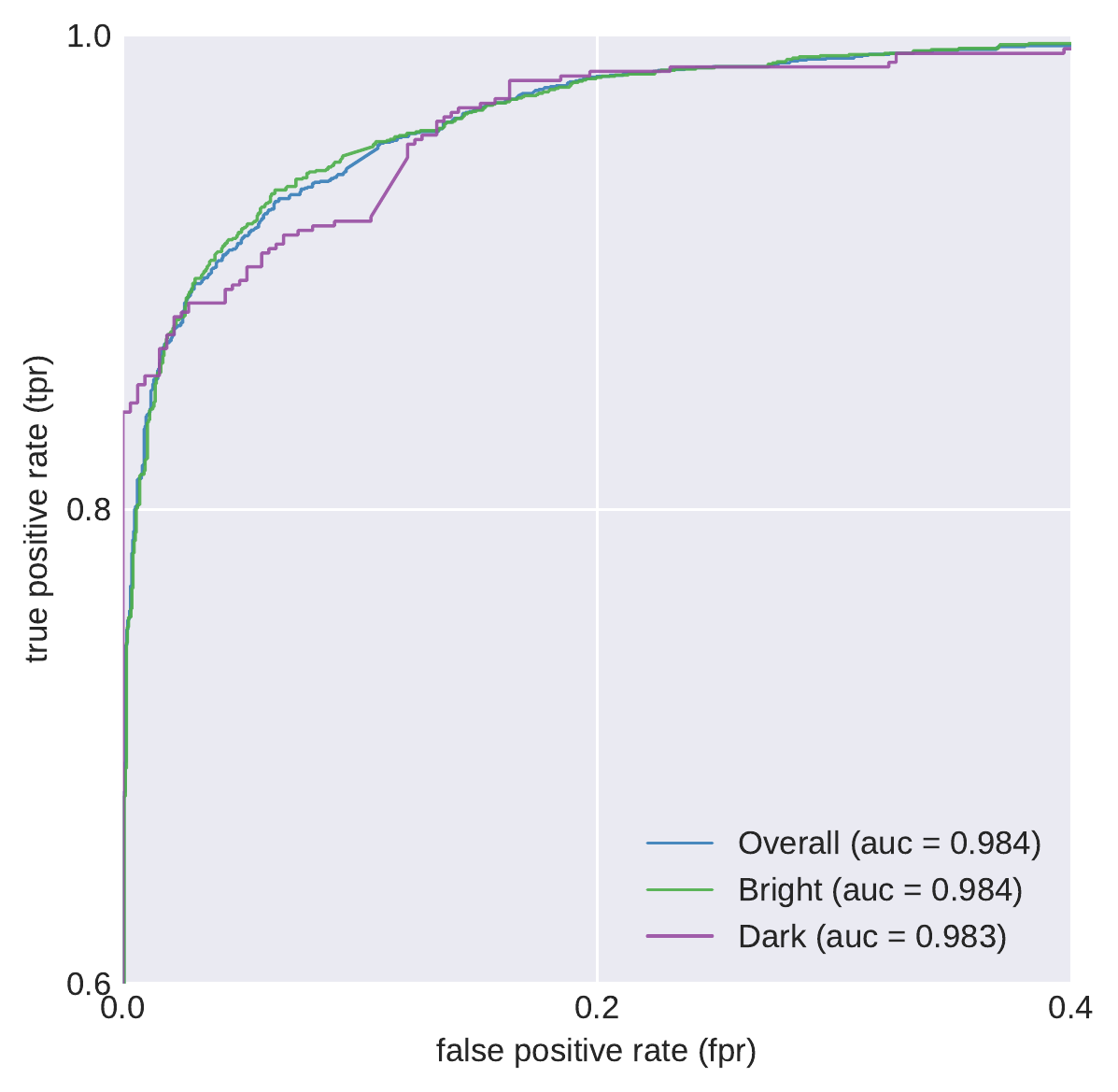}
    \end{subfigure}
    \\
    \rotatebox[origin=l]{90}{\small WM~\cite{WM2020}}
    &
    \begin{subfigure}{\scaleroc\linewidth}
        \includegraphics[width=\textwidth]{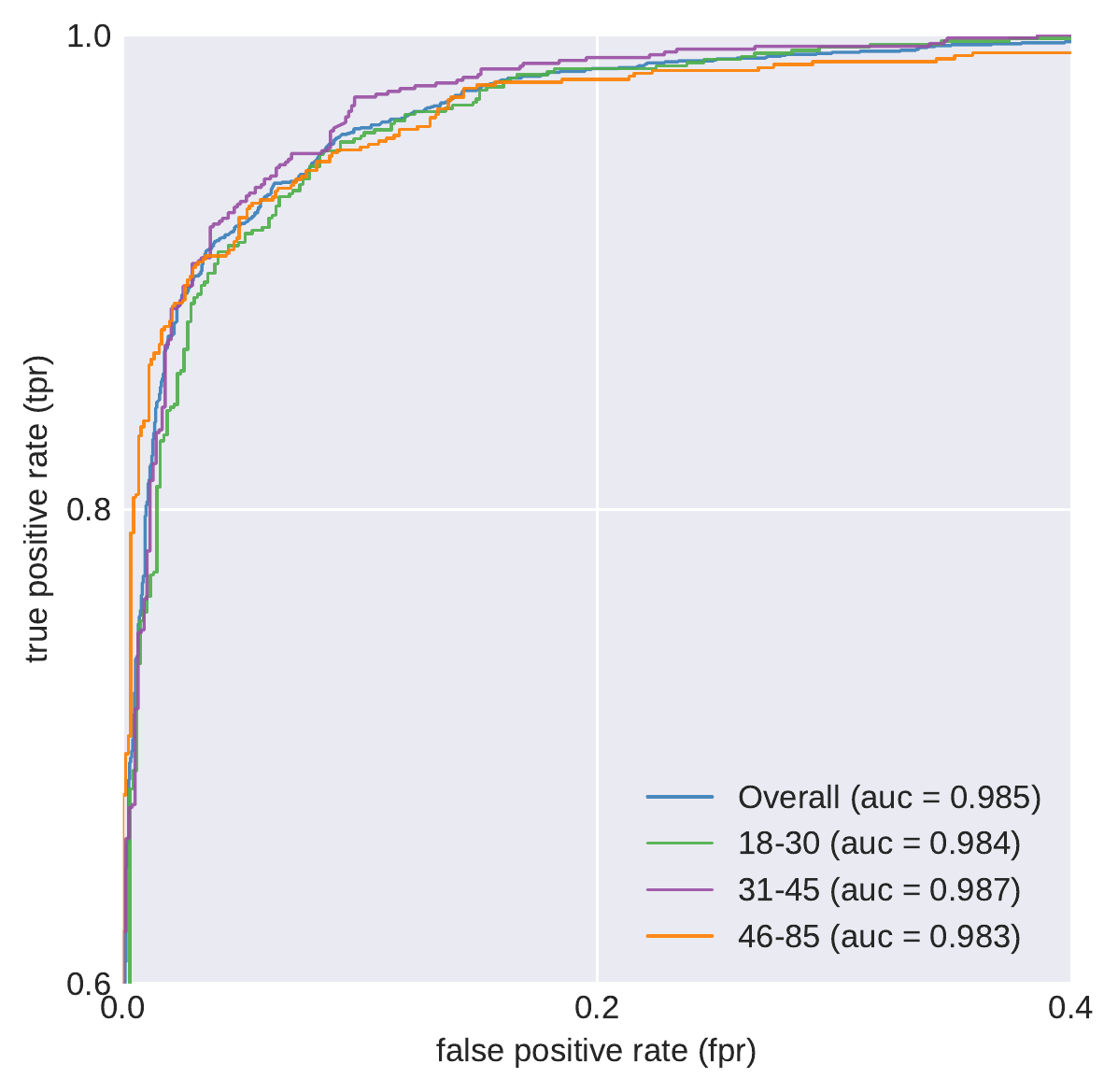}
    \end{subfigure}
    &
    \begin{subfigure}{\scaleroc\linewidth}
        \includegraphics[width=\textwidth]{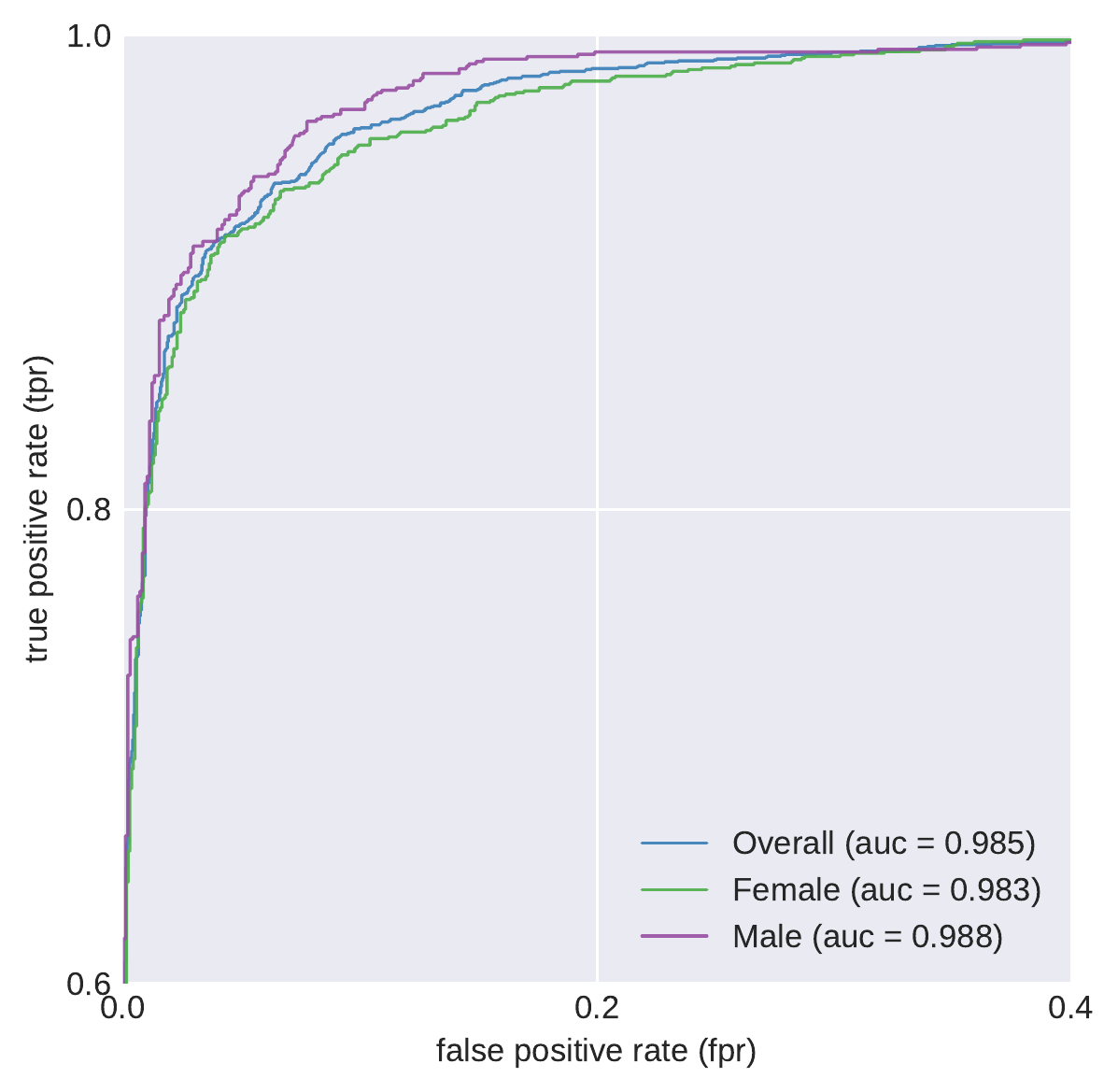}
    \end{subfigure}
    &
    \begin{subfigure}{\scaleroc\linewidth}
        \includegraphics[width=\textwidth]{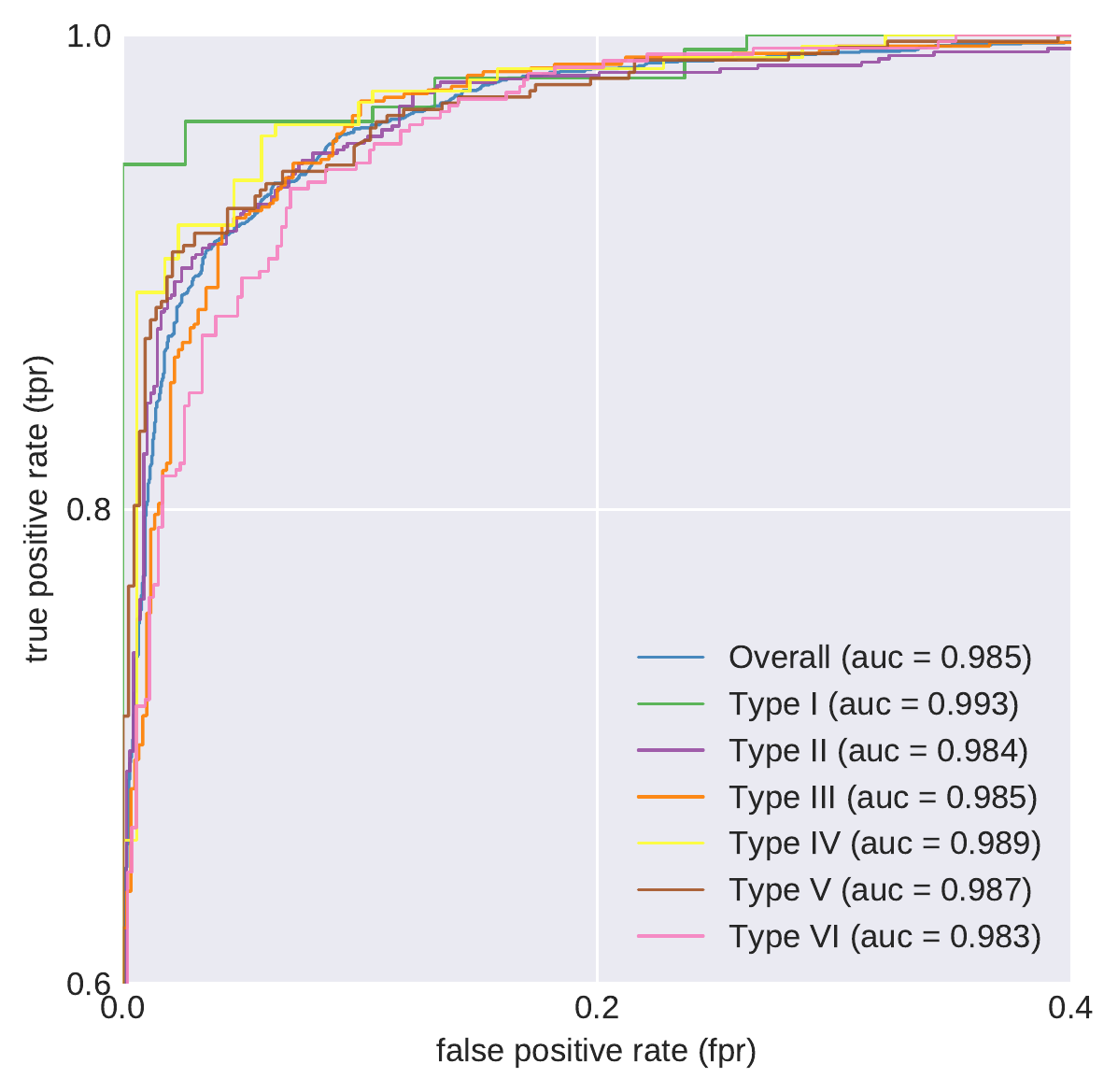}
    \end{subfigure}
    &
    \begin{subfigure}{\scaleroc\linewidth}
        \includegraphics[width=\textwidth]{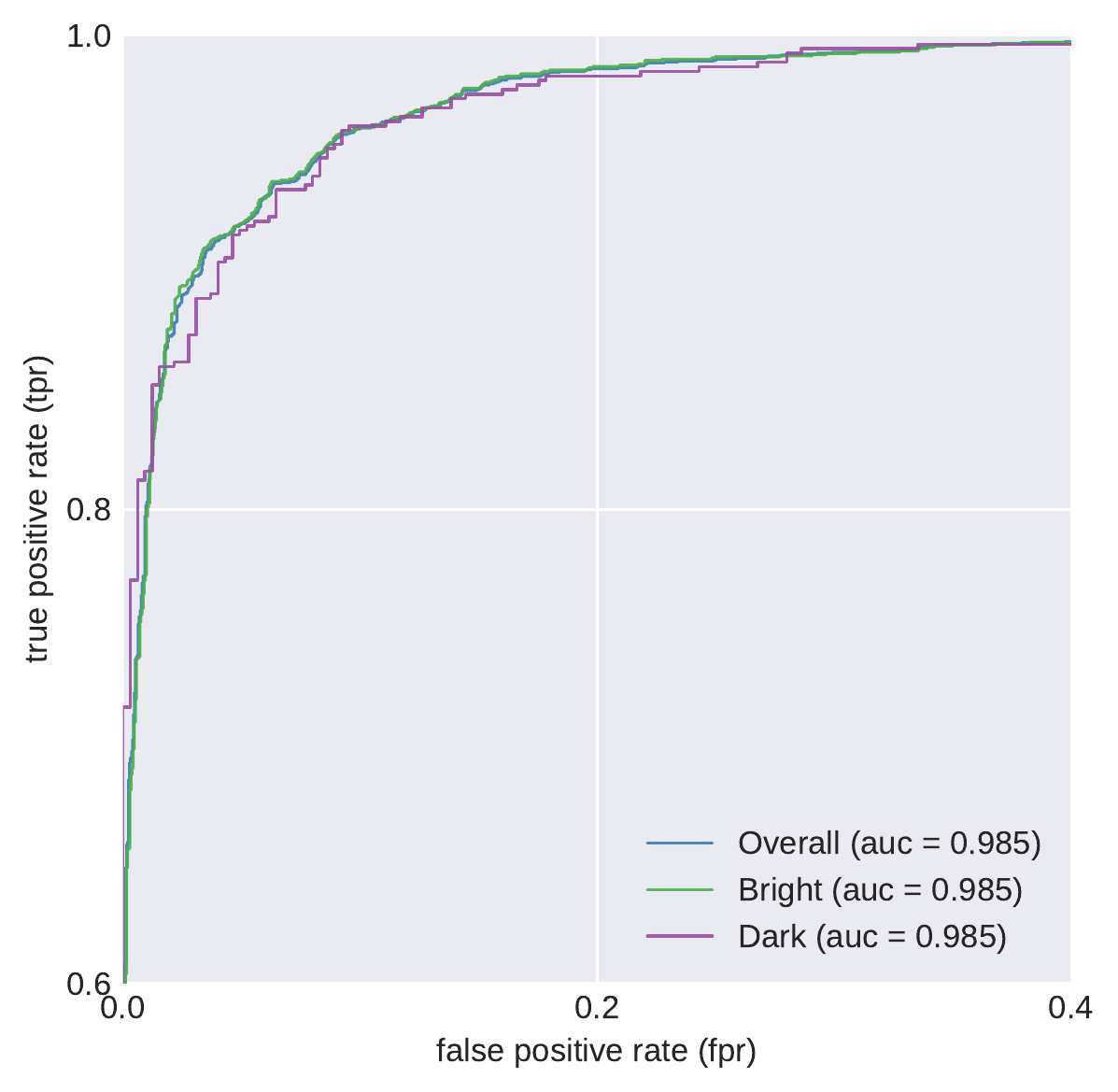}
    \end{subfigure}
    \\
    \rotatebox[origin=l]{90}{\small NTechLab~\cite{NTechLab2020}}
    &
    \begin{subfigure}{\scaleroc\linewidth}
        \includegraphics[width=\textwidth]{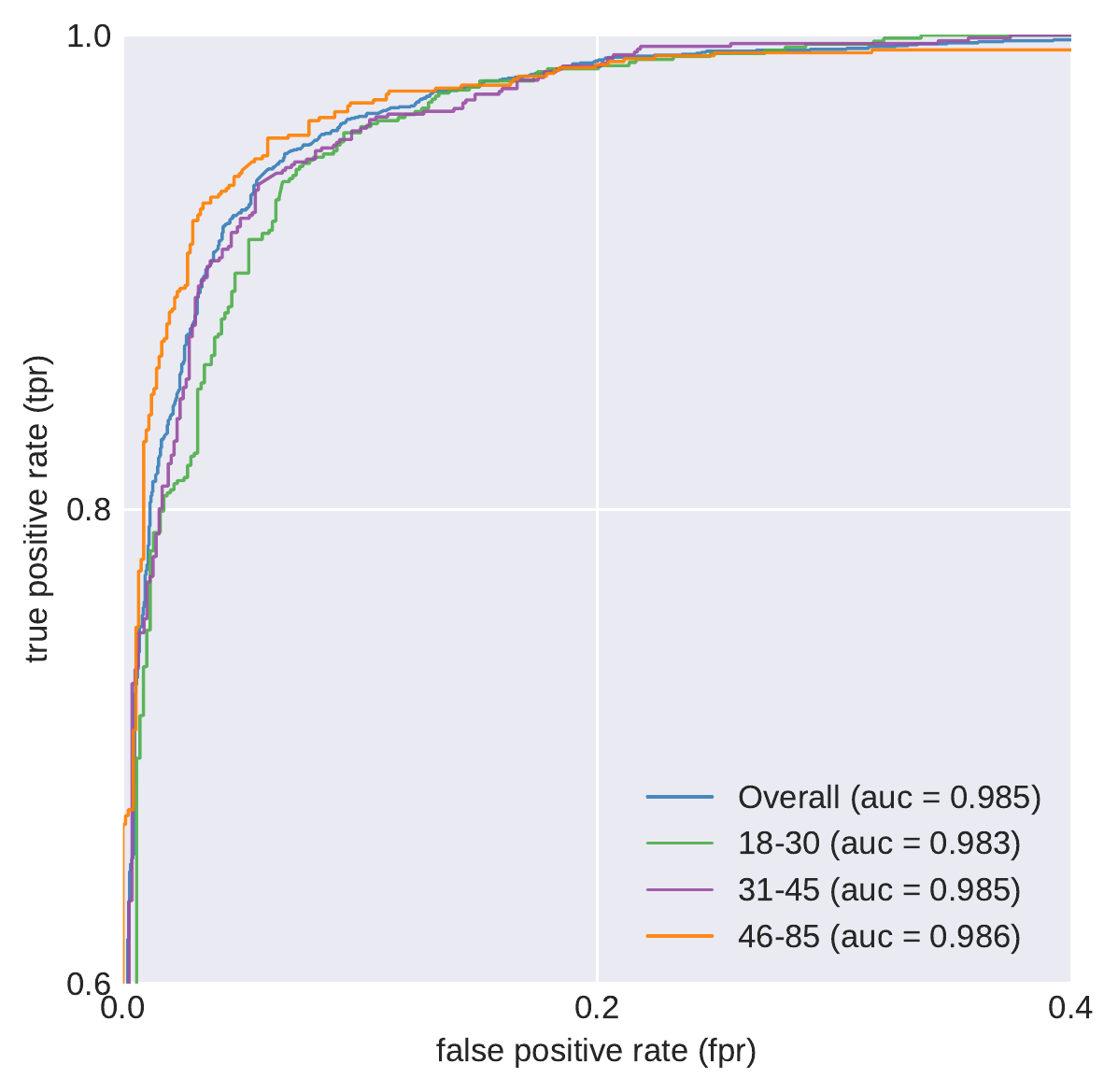}
    \end{subfigure}
    &
    \begin{subfigure}{\scaleroc\linewidth}
        \includegraphics[width=\textwidth]{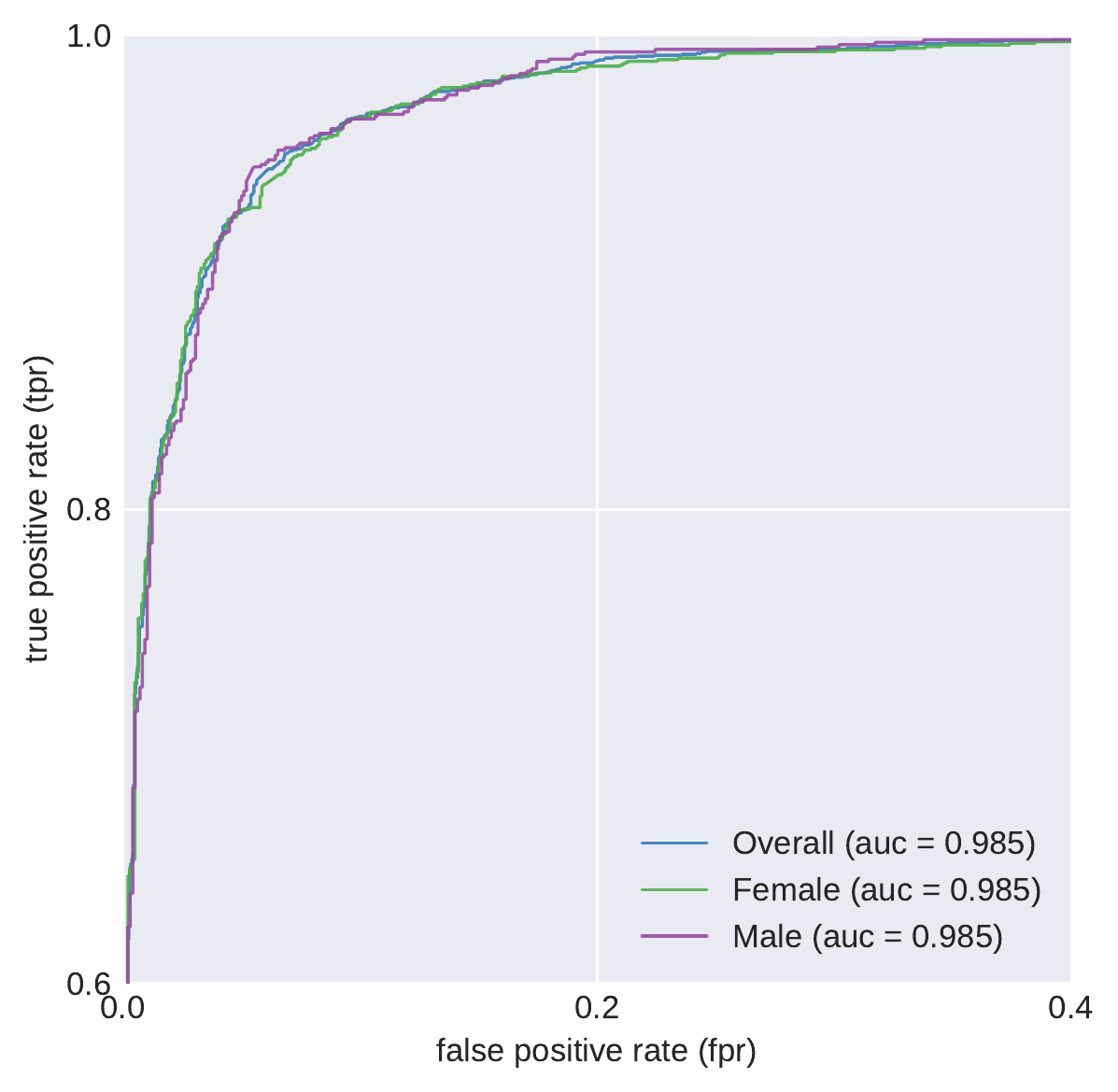}
    \end{subfigure}
    &
    \begin{subfigure}{\scaleroc\linewidth}
        \includegraphics[width=\textwidth]{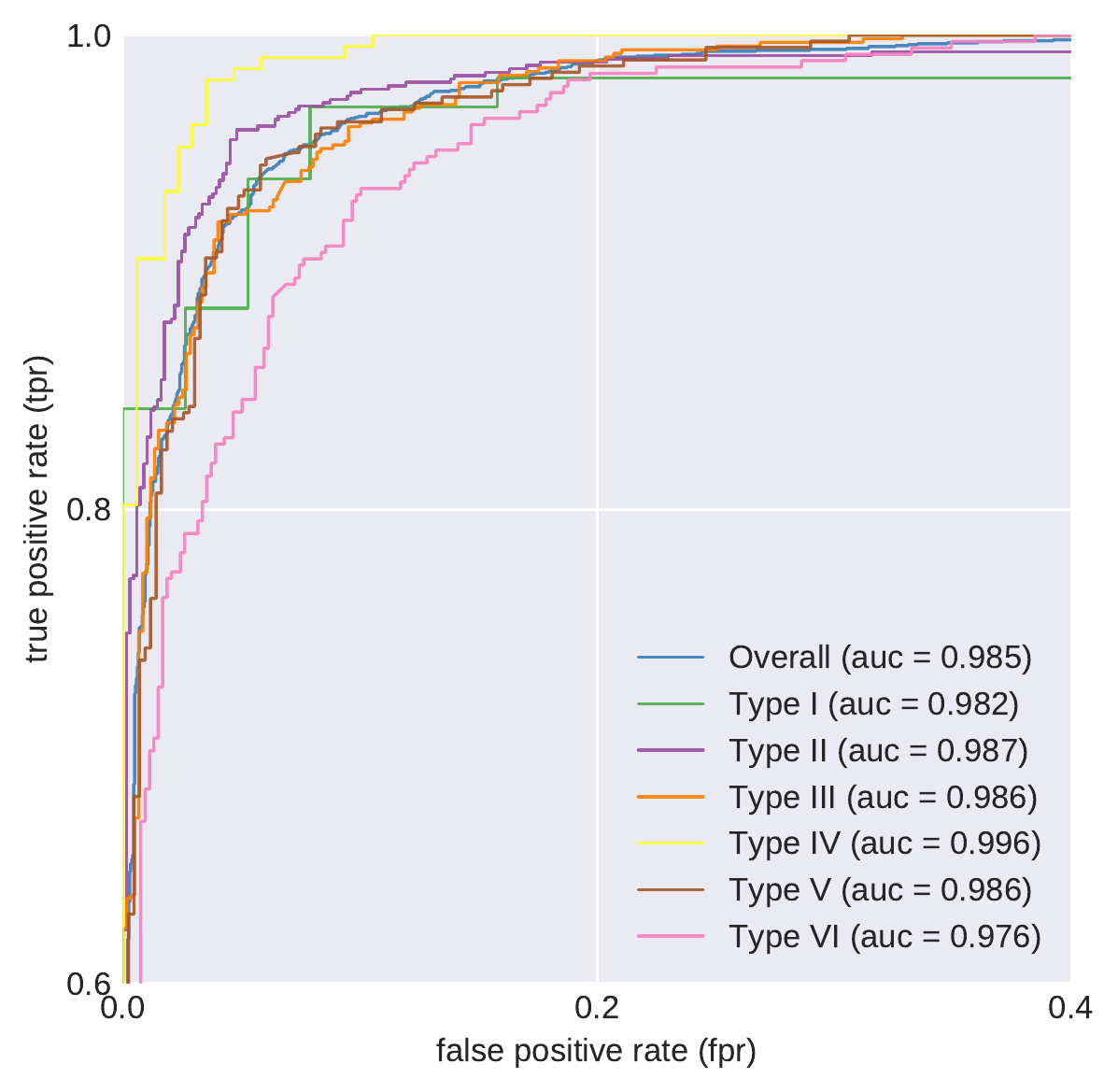}
    \end{subfigure}
    &
    \begin{subfigure}{\scaleroc\linewidth}
        \includegraphics[width=\textwidth]{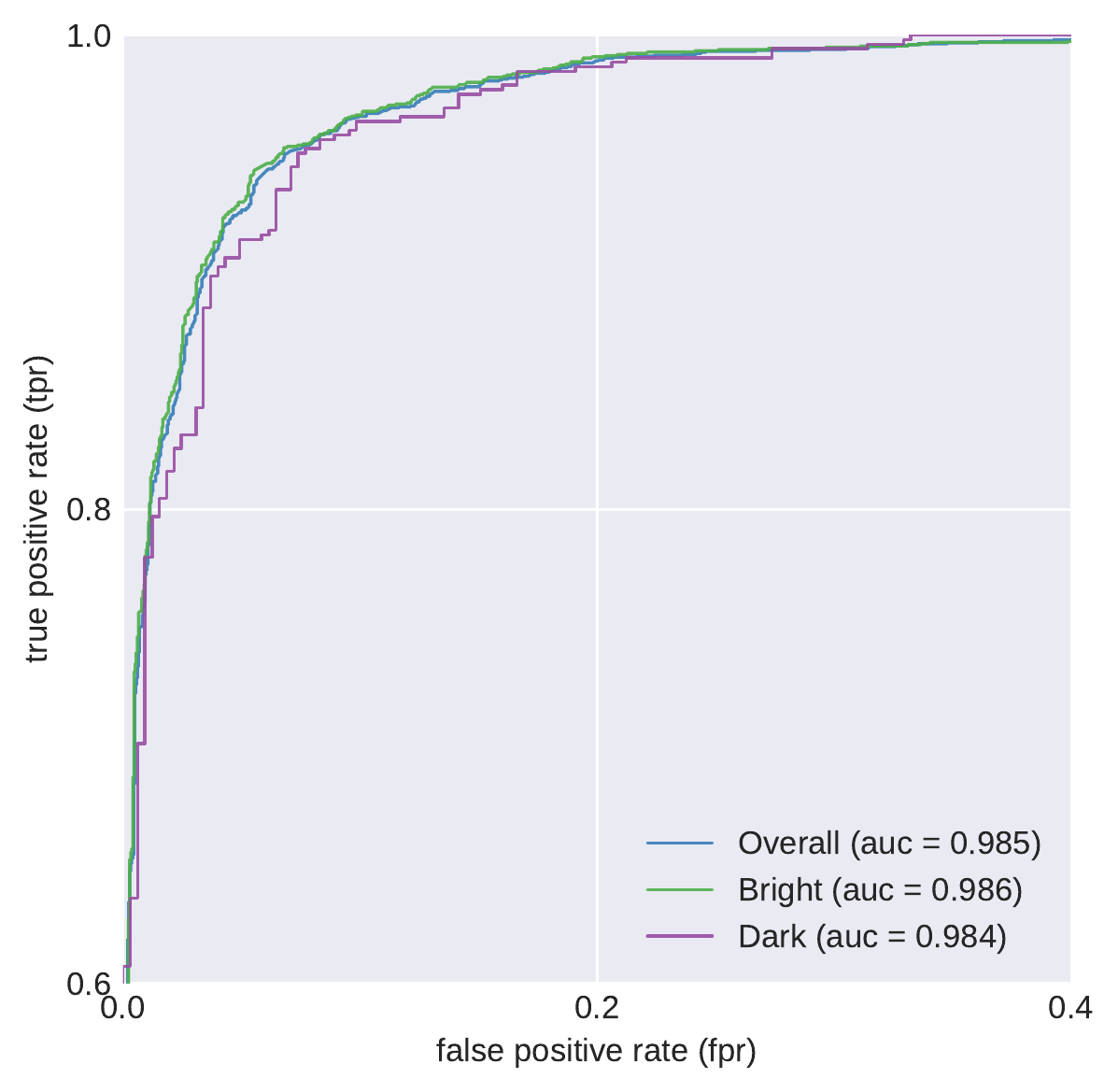}
    \end{subfigure}
    \\
    \rotatebox[origin=l]{90}{\small Eighteen Years Old~\cite{Eighteen2020}}
    &
    \begin{subfigure}{\scaleroc\linewidth}
        \includegraphics[width=\textwidth]{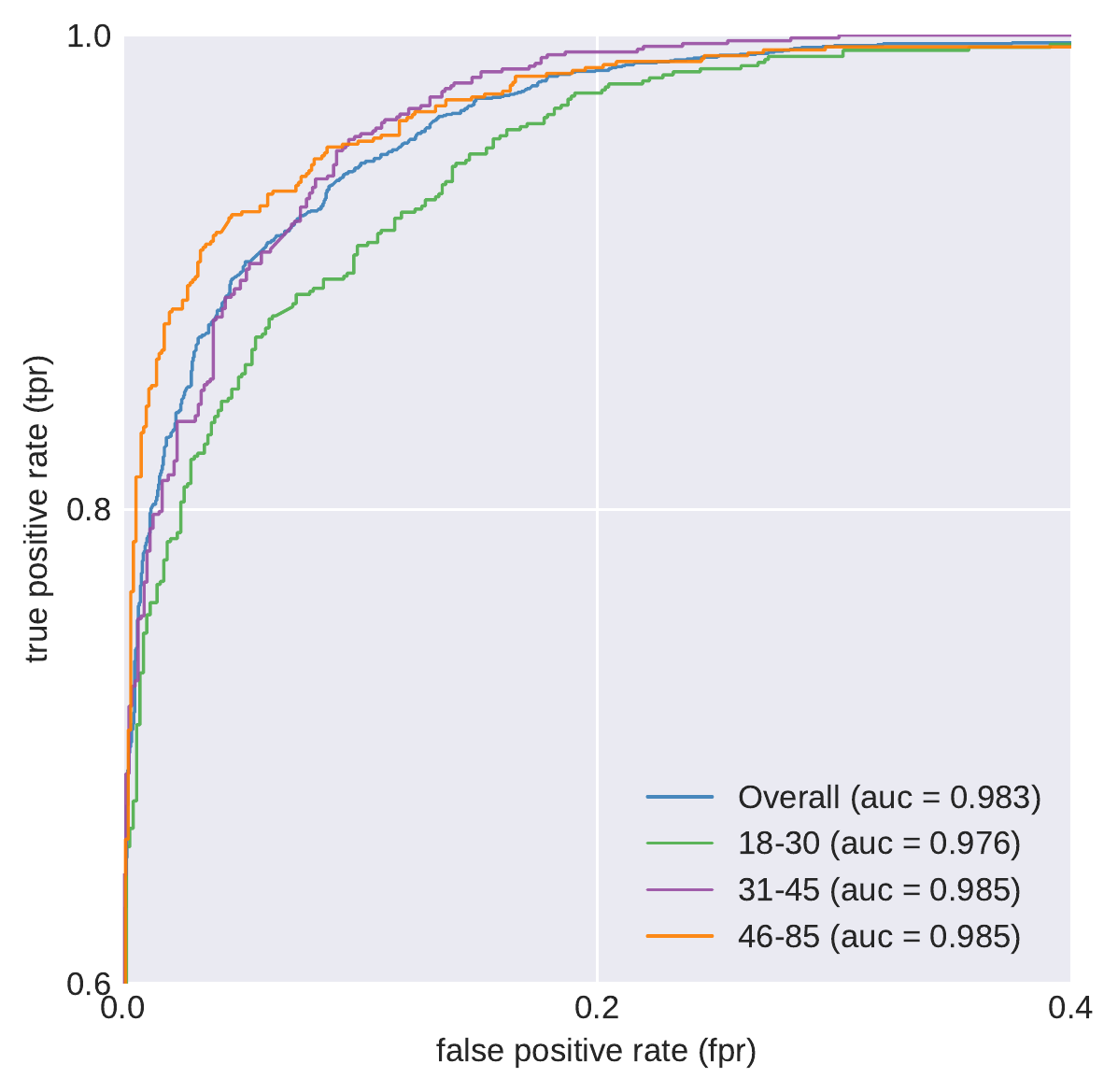}
    \end{subfigure}
    &
    \begin{subfigure}{\scaleroc\linewidth}
        \includegraphics[width=\textwidth]{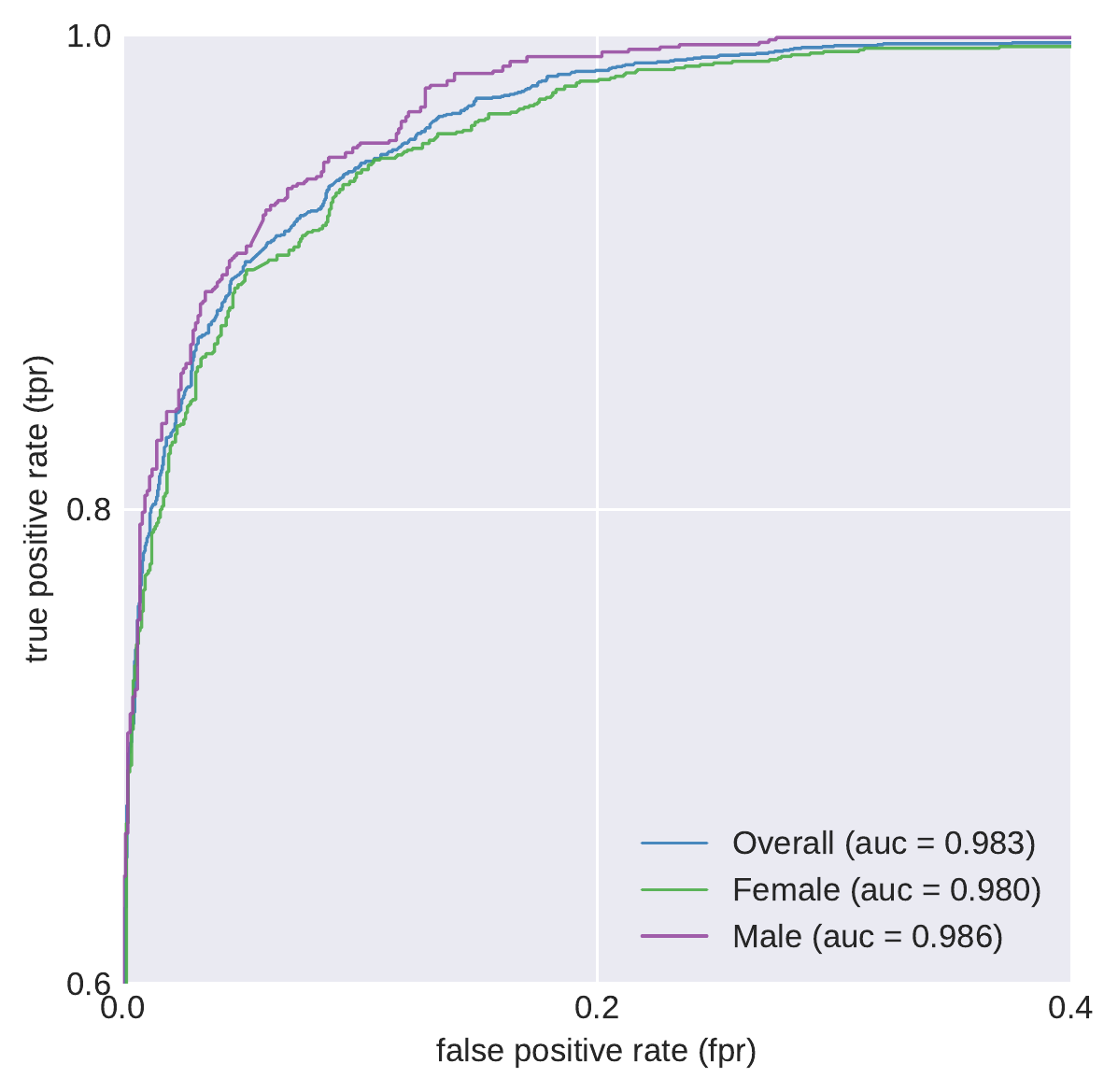}
    \end{subfigure}
    &
    \begin{subfigure}{\scaleroc\linewidth}
        \includegraphics[width=\textwidth]{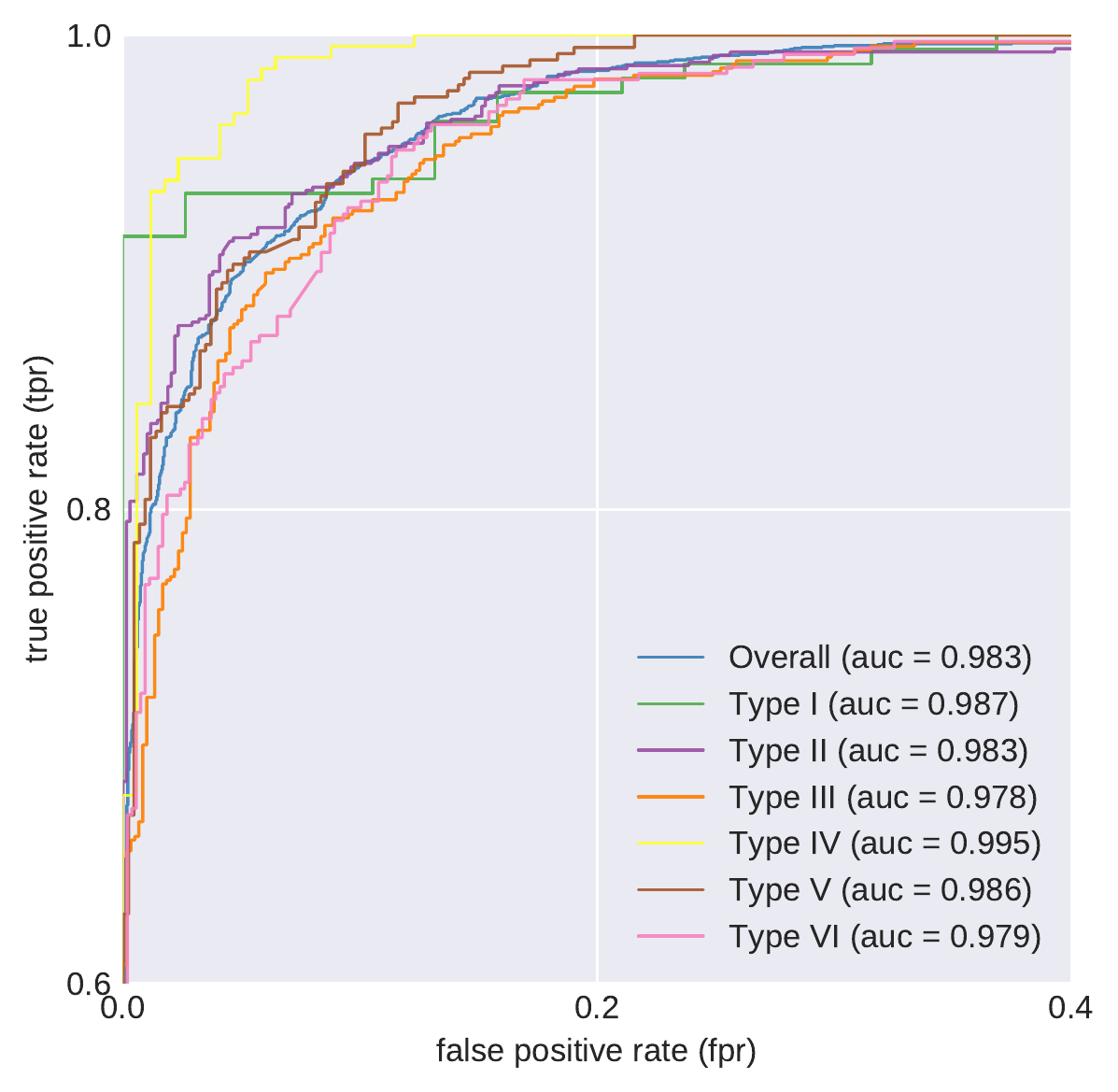}
    \end{subfigure}
    &
    \begin{subfigure}{\scaleroc\linewidth}
        \includegraphics[width=\textwidth]{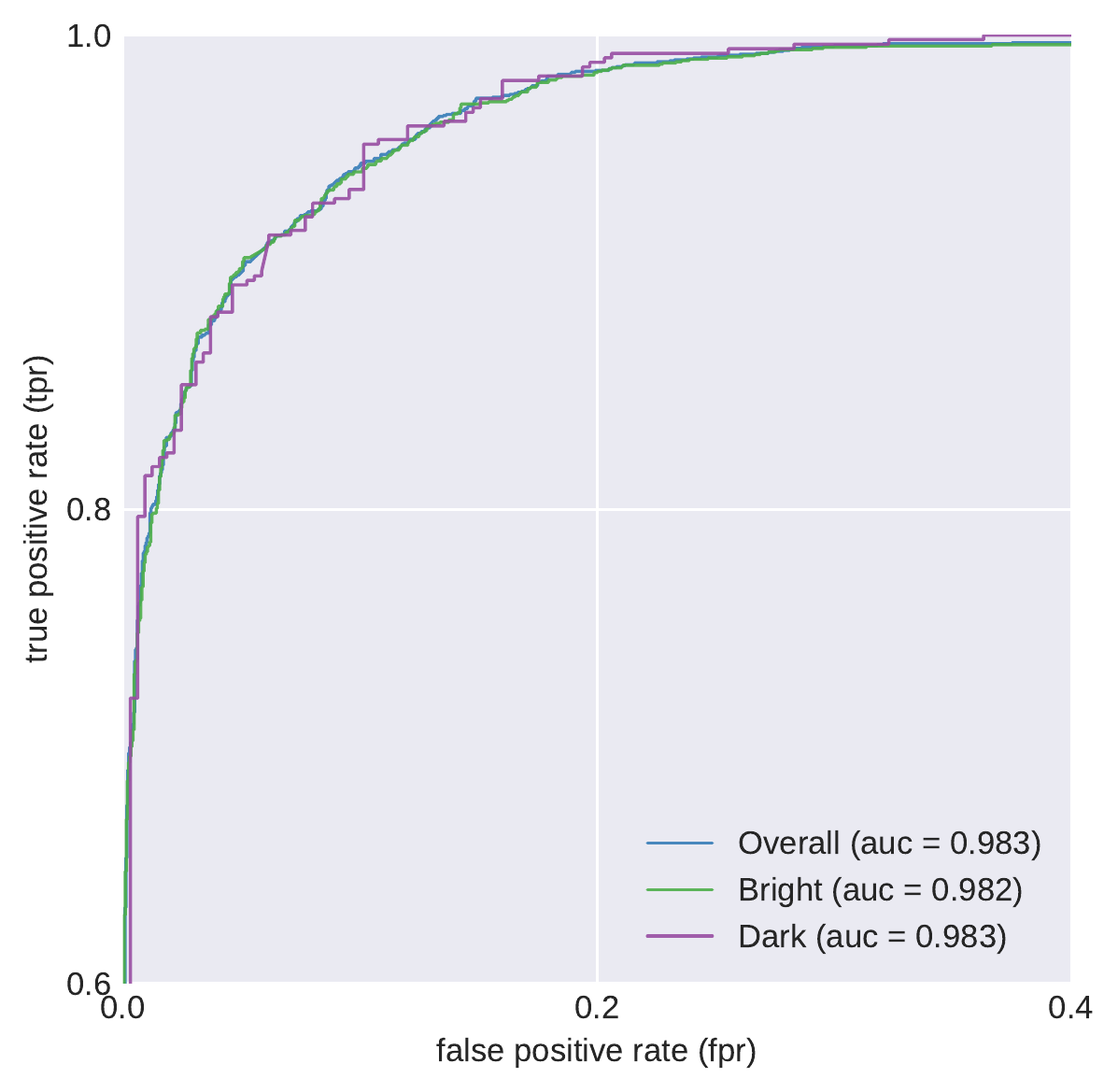}
    \end{subfigure}
    \\
    \rotatebox[origin=l]{90}{\small The Medics~\cite{Medics2020}}
    &
    \begin{subfigure}{\scaleroc\linewidth}
        \includegraphics[width=\textwidth]{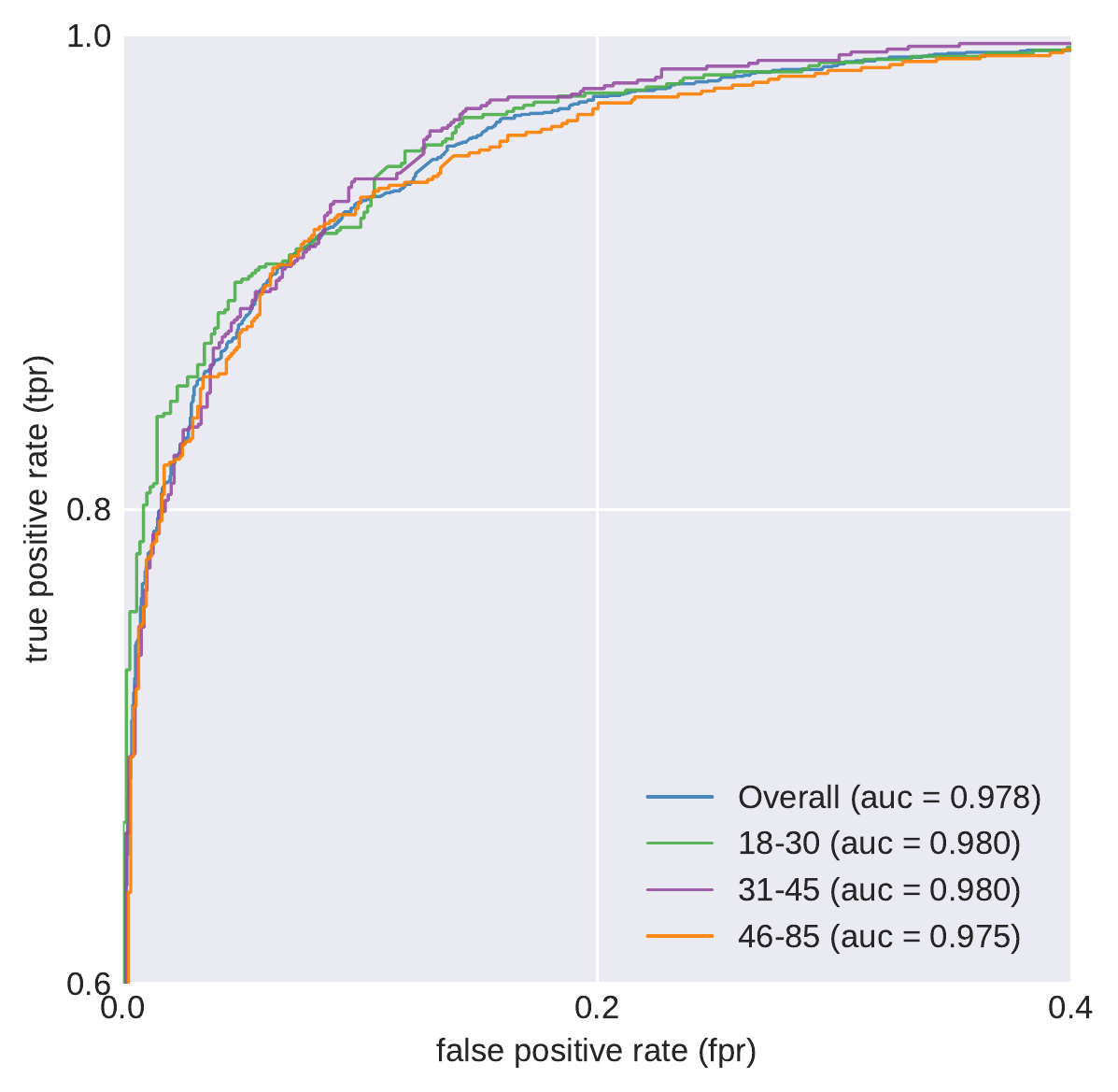}
    \end{subfigure}
    &
    \begin{subfigure}{\scaleroc\linewidth}
        \includegraphics[width=\textwidth]{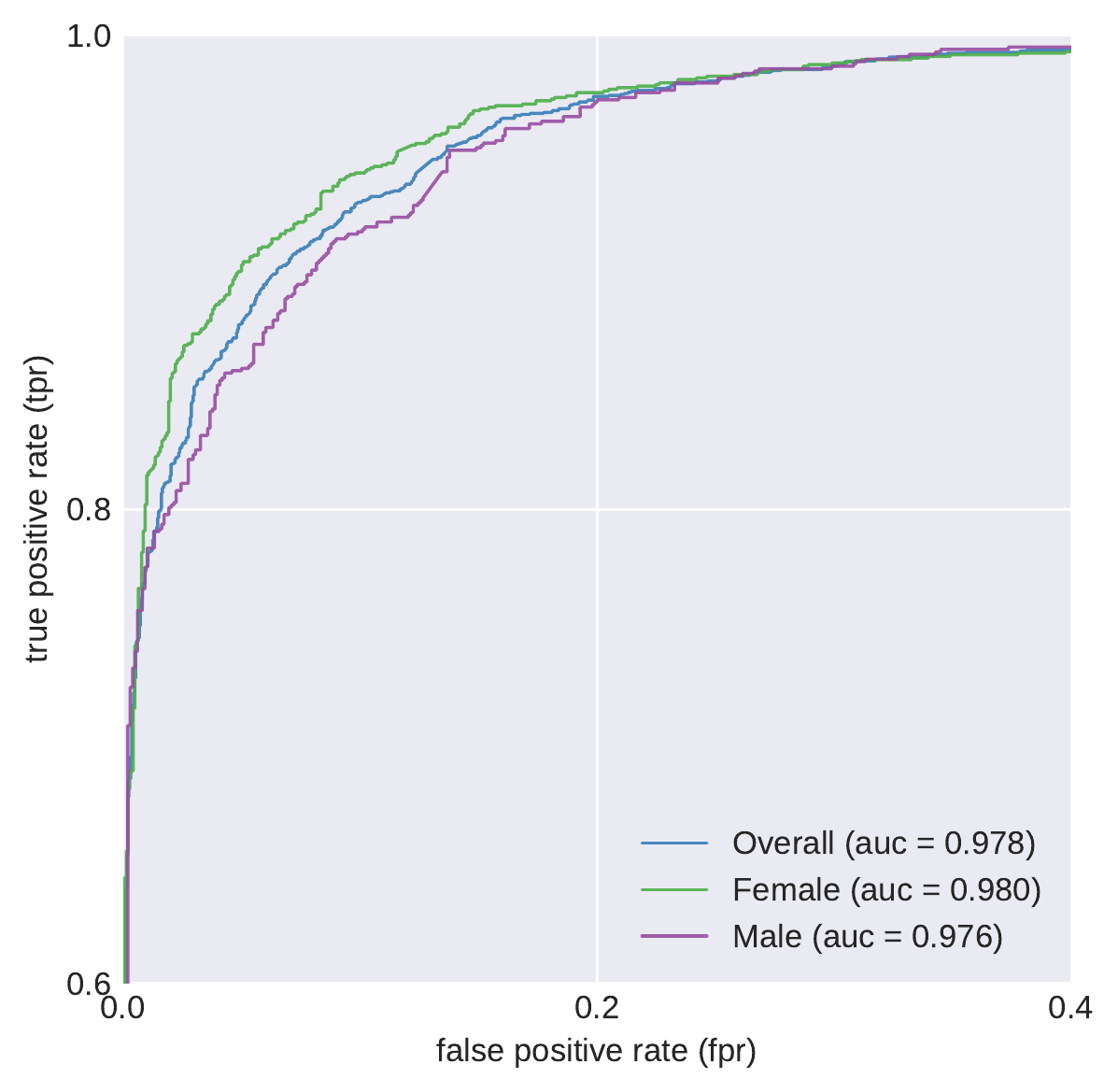}
    \end{subfigure}
    &
    \begin{subfigure}{\scaleroc\linewidth}
        \includegraphics[width=\textwidth]{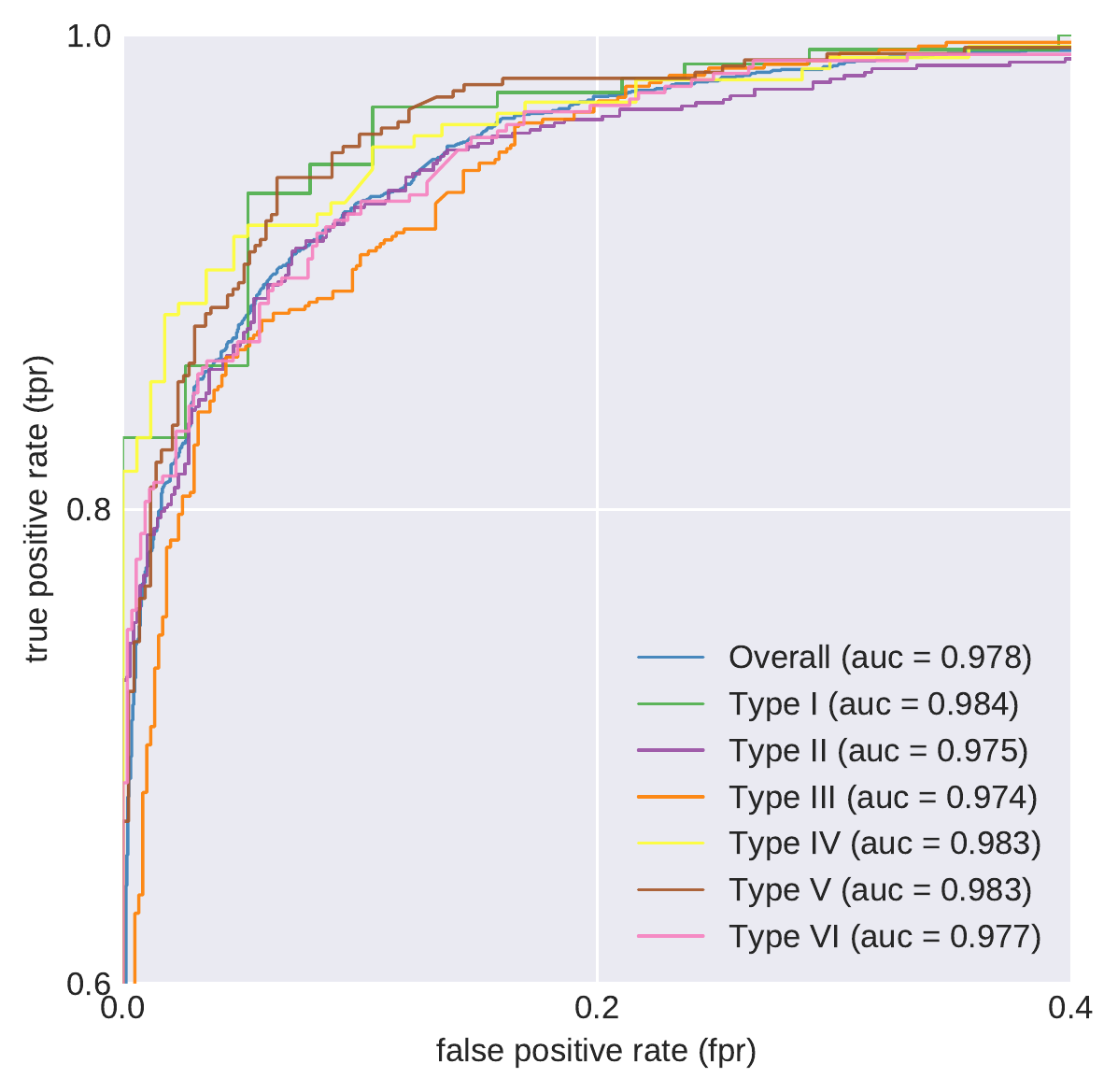}
    \end{subfigure}
    &
    \begin{subfigure}{\scaleroc\linewidth}
        \includegraphics[width=\textwidth]{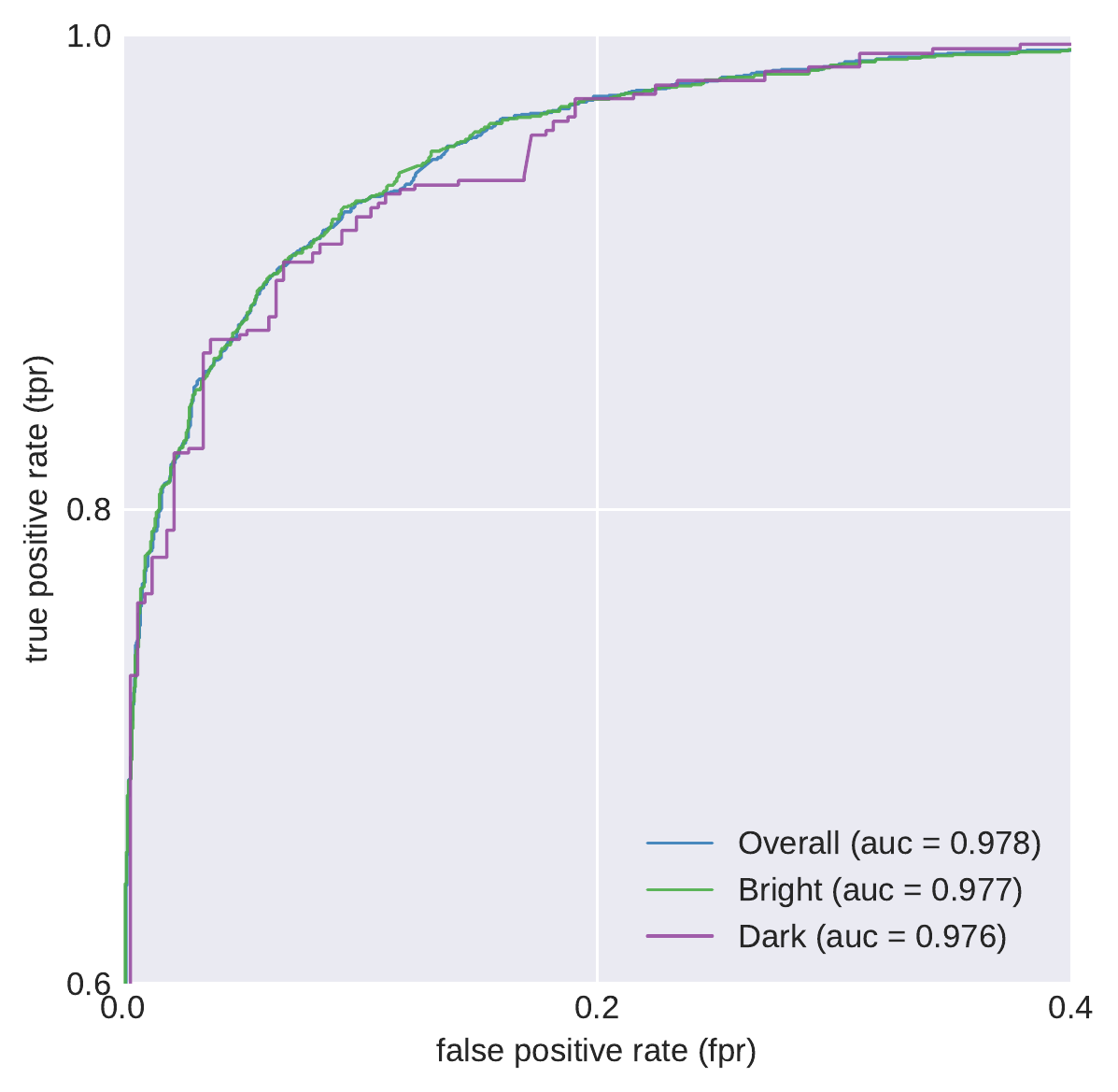}
    \end{subfigure}
    \\
    \arrayrulecolor{lightgray}\cmidrule{2-5}
    \end{tabular}
    \caption{\label{fig:dfdc_roc_curves} \textbf{DFDC top winner's} ROC curves, breakdown on the Fairness annotations. Selim Seferbekov~\cite{Seferbekov2020} (top winner of the DFDC) has similar ROC curves for age groups but it is more accurate on \textit{Female} examples and more sensitive to pale or very dark skin tones (\textit{Type I} and \textit{Type VI}). WM~\cite{WM2020} generally preserves the balance across all categories except the skin type. On the other hand, NTechLab~\cite{NTechLab2020} curves are more class-balanced over age, gender and lighting, while Eighteen Years Old~\cite{Eighteen2020} is invariant to lighting. The Medics~\cite{Medics2020}, on the other hand, does not preserve balance in most of the categories. Axes of ROC curves are zoomed-in for better visualization.}
\end{figure*}

\section{\label{sec:conclusion}Conclusions}
We presented the \textit{Casual Conversations Dataset}, a dataset designed to measure robustness of AI models across four main dimensions, \textit{age}, \textit{gender}, \textit{apparent skin type} and \textit{lighting}. As previously stated, a unique factor of our dataset is that the \textit{age} and \textit{gender} labels are provided by the participants themselves. The dataset has uniform distributions over all categories and could be used to measure various AI methods, such as face detection, apparent age and gender classification, or to assess robustness to various ambient lighting conditions.

As an application of our dataset, we presented an analysis of the DeepFake Detection Challenge's top five winners on the four main dimensions of the dataset. We presented the precision and log-loss scores of winners on the intersection of the \textit{DFDC} private test set and the \textit{Casual Conversations} dataset. From our thorough investigation, we conclude that all methods carry a large bias towards lighter skin tones since they mostly fail on darker-skinned subjects. 

Moreover, we also discussed the results of the recent apparent age and gender prediction models on our dataset. In both of the applications, we noticed an obvious algorithmic bias towards lighter-skinned subjects. Apparent gender classification methods are most successful on older people ($+45$ years old) and generally as good on darker videos as on brighter ones.

Beyond aforementioned research topics, our dataset enables researchers to develop and also thoroughly evaluate models for more inclusive and responsible AI.

\ifCLASSOPTIONcompsoc
  \section*{Acknowledgments}
\else
  \section*{Acknowledgment}
\fi
We would like to thank Ida Cheng and Tashrima Hossain for their help in regards to annotating the dataset for the Fitzpatrick skin type.

\ifCLASSOPTIONcaptionsoff
  \newpage
\fi

\bibliographystyle{IEEEtran}
\bibliography{bibliography}

%
\vspace{-15pt}
\begin{IEEEbiography}[{\includegraphics[width=1in,height=1.25in,clip,keepaspectratio]{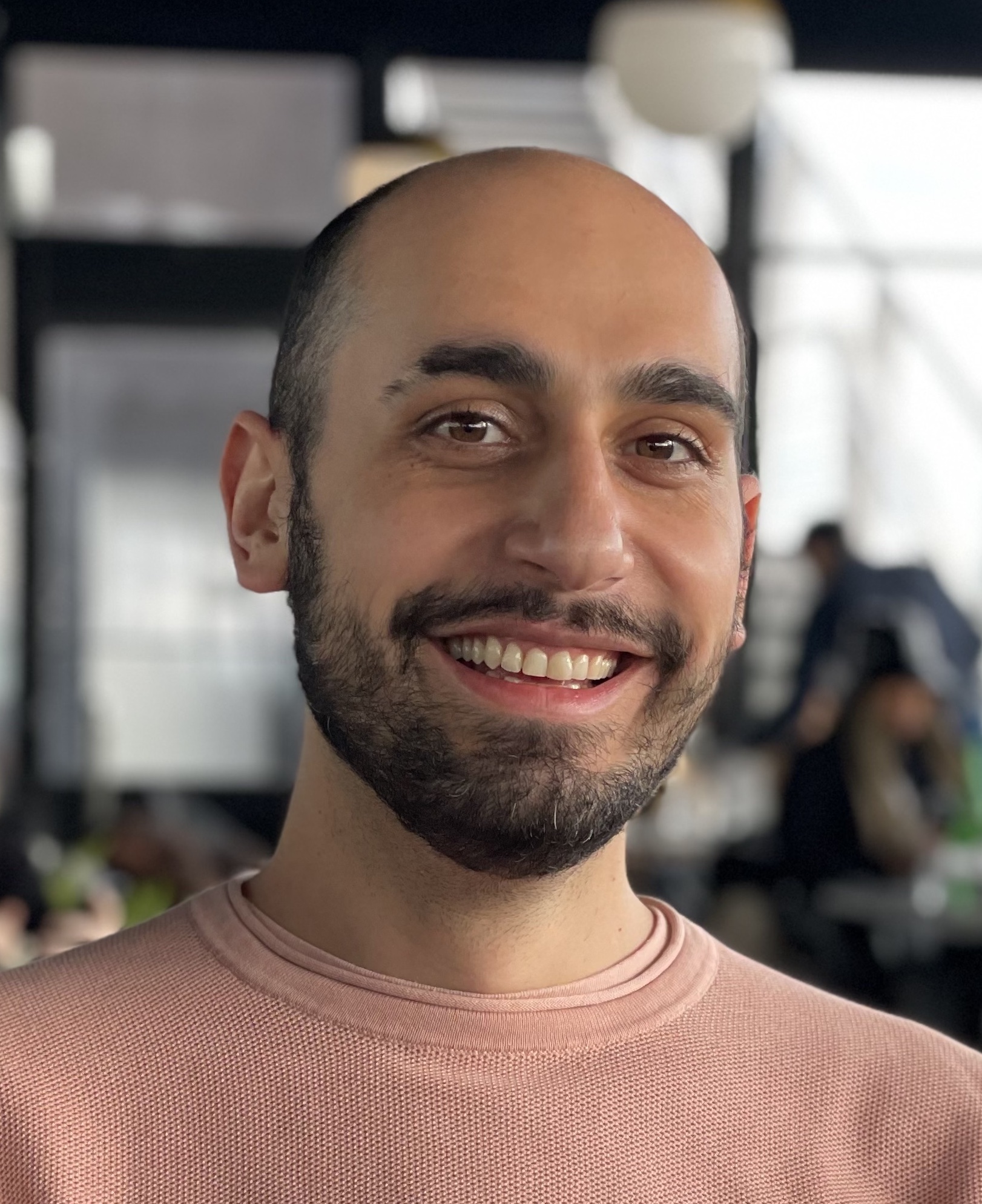}}]{Caner Hazirbas} is a research scientist in the AI Red Team at Facebook AI. Previously, he worked at Apple as a deep learning engineer (USA). He received his PhD and MS from the Technical University of Munich (Germany). His research interests lie in the field of computer vision, adversarial machine learning and fairness in AI.
\end{IEEEbiography}
\vspace{-15pt}

\begin{IEEEbiography}[{\includegraphics[width=1in,height=1.25in,clip,keepaspectratio]{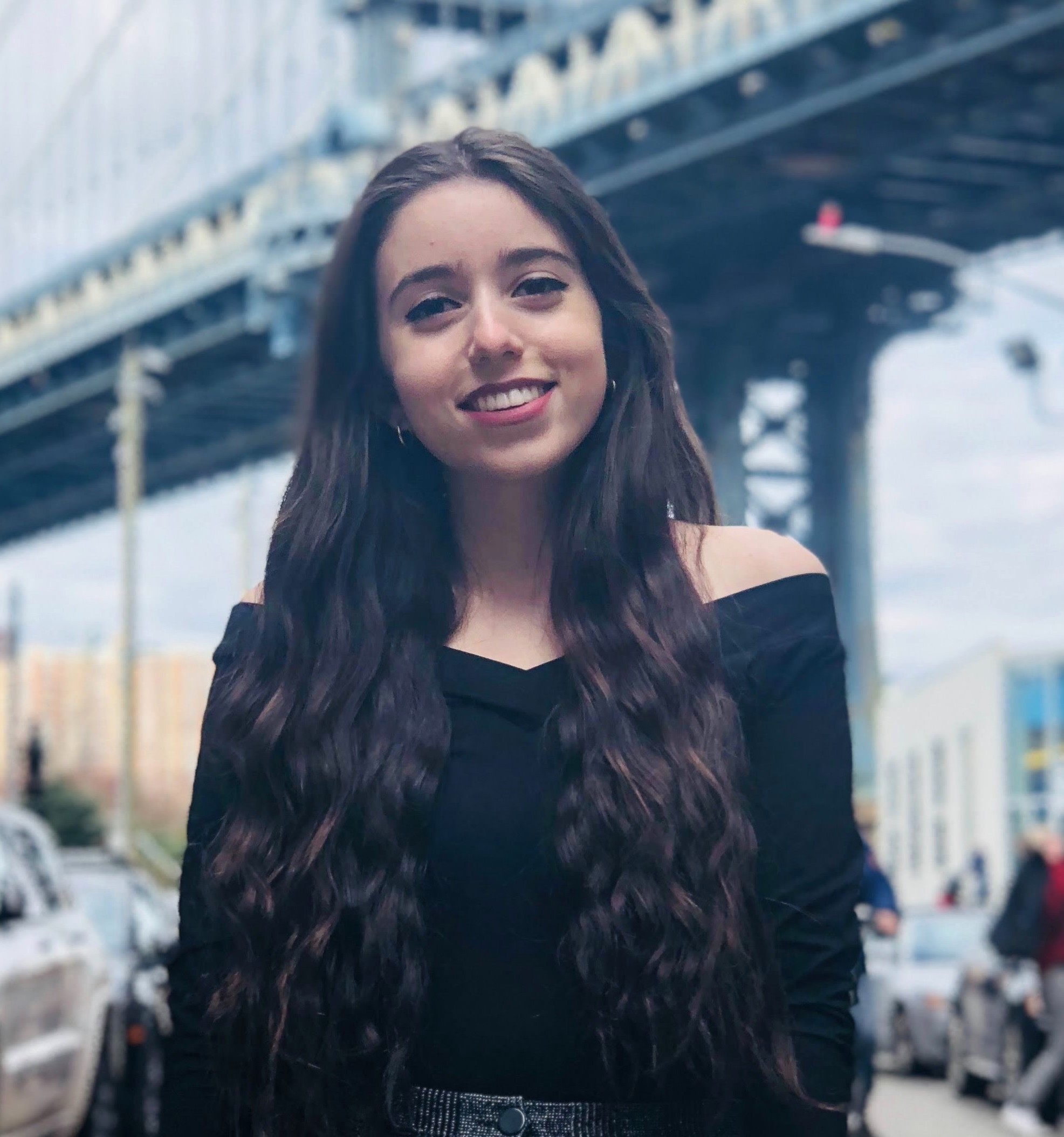}}]{Joanna Bitton} is a software engineer in the AI Red Team at Facebook AI. She received her Masters in Computer Science from the Courant Institute at New York University. Her research interests lie in the field of computer vision, adversarial machine learning, and fairness in AI.
\end{IEEEbiography}
\vspace{-15pt}

\begin{IEEEbiographynophoto}{Brian Dolhansky} is a machine learning staff engineer at Reddit. He was earlier a research scientist on the AI Red Team within the Facebook Applied AI Research (FAIAR) group. Previously he was a graduate student at the University of Washington studying theoretical and applied machine learning. 
\end{IEEEbiographynophoto}
\vspace{-15pt}

\begin{IEEEbiography}[{\includegraphics[width=1in,height=1.25in,clip,keepaspectratio]{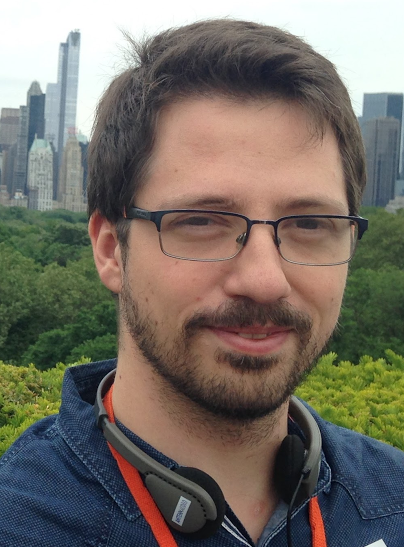}}]{Albert Gordo} is a research scientist in the AI Red Team at Facebook AI. He received his PhD from the Computer Vision Center in the Universitat Autonoma de Barcelona (Spain), in collaboration with the Computer Vision group at XRCE. After that he was a postdoc at the LEAR group in INRIA Grenoble (France), working on large-scale object detection. From 2014 to 2017 he was a research scientist in the Computer Vision group at XRCE.
\end{IEEEbiography}
\vspace{-15pt}

\begin{IEEEbiography}[{\includegraphics[width=1in,height=1.25in,clip,keepaspectratio]{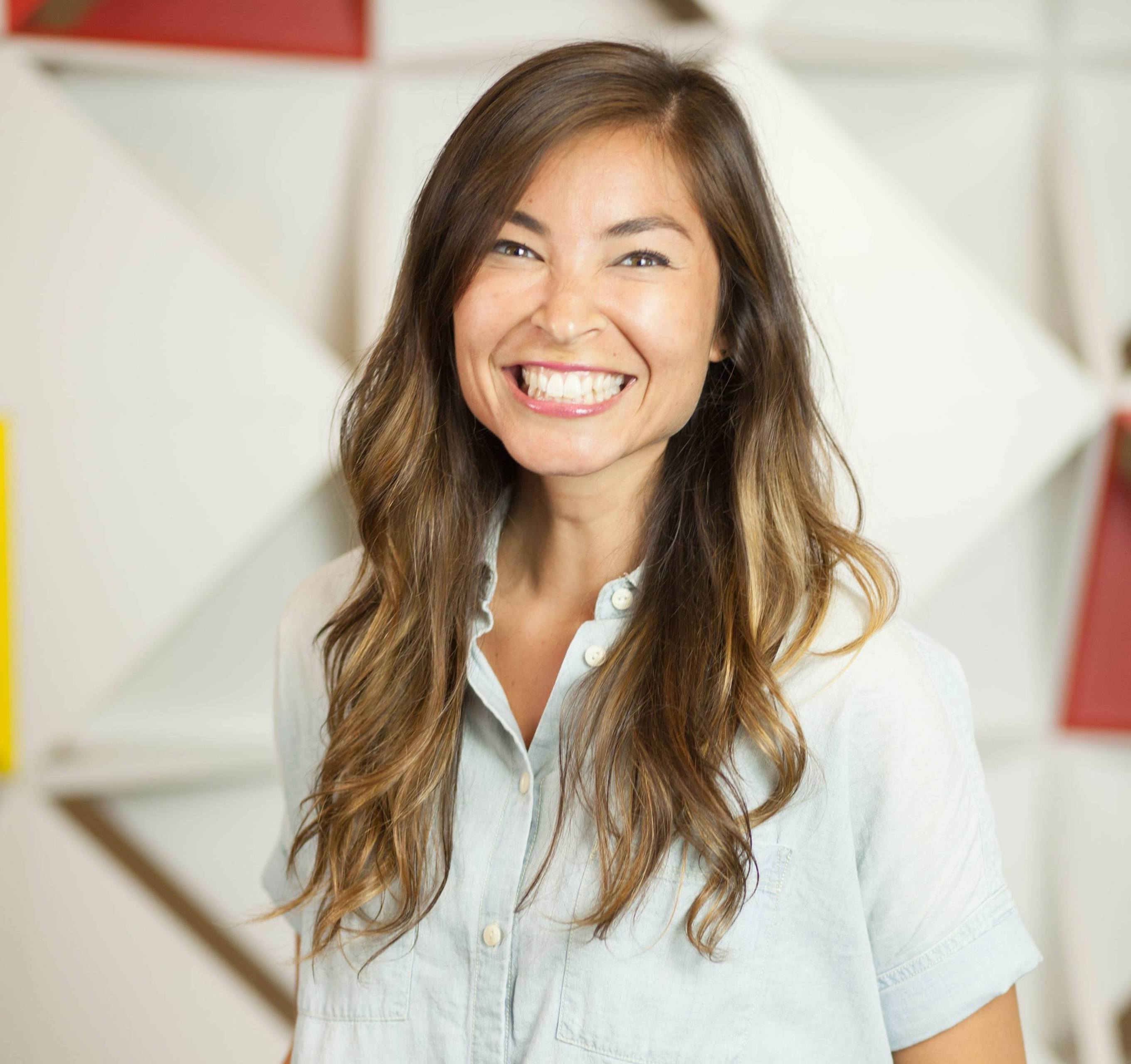}}]{Jacqueline Pan} is a program manager at Facebook AI and is focused on developing Responsible AI practices and programs across the company. Prior to Facebook, Jacqueline was the program lead for ML Fairness at Google and led a team that developed programs to understand and improve inclusivity in machine learning.
\end{IEEEbiography}
\vspace{-15pt}

\begin{IEEEbiography}[{\includegraphics[width=1in,height=1.25in,clip,keepaspectratio]{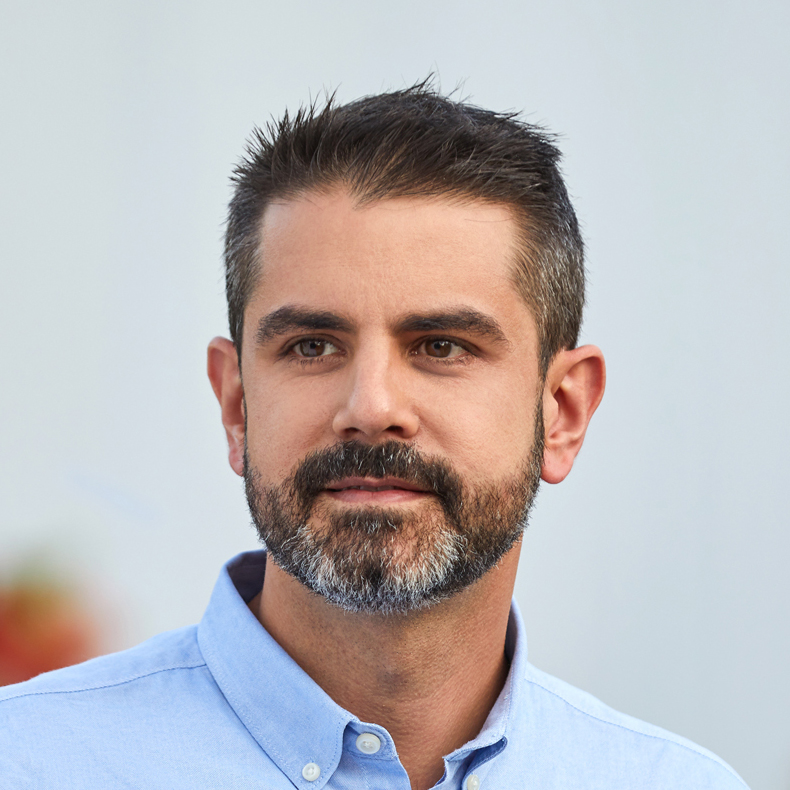}}]{Cristian Canton Ferrer} is a research manager at Facebook AI, focusing on understanding weaknesses and vulnerabilities derived from the use (or misuse) of AI. In the past, he managed the computer vision team within the objectionable and harmful content domain. From 2012-16, he was at Microsoft Research in Redmond (USA) and Cambridge (UK); from 2009-2012, he was at Vicon (Oxford), bringing computer vision to produce visual effects for the cinema industry. He got his PhD and MS from Technical University of Catalonia (Barcelona) and his MS Thesis from EPFL (Switzerland) on computer vision topics.
\end{IEEEbiography}
\vfill

\end{document}